\documentclass[10pt,twocolumn,letterpaper]{article}

\usepackage{cvpr} 

\usepackage{graphicx}
\usepackage{amsmath}
\usepackage{amssymb}
\usepackage{booktabs}
\usepackage{subfiles}
\usepackage{multirow}
\usepackage{stmaryrd}
\usepackage{soul}
\usepackage[dvipsnames]{xcolor}
\usepackage{tabularx}
\usepackage{bbding}
\usepackage{algorithm}
\usepackage{algpseudocode}
\usepackage{rotating}
\usepackage{longtable}
\graphicspath{ {./images/} }

\usepackage[pagebackref,breaklinks,colorlinks]{hyperref}

\usepackage[capitalize]{cleveref}
\crefname{section}{Sec.}{Secs.}
\Crefname{section}{Section}{Sections}
\Crefname{table}{Table}{Tables}
\crefname{table}{Tab.}{Tabs.}

\def\paperID{15375} % *** Enter the Paper ID here
\def\confName{CVPR}
\def\confYear{2024}
\title{Supervised Anomaly Detection for Complex Industrial Images}

\author{
Aimira Baitieva$^1$, David Hurych$^2$, Victor Besnier$^2$, Olivier Bernard$^1$\\
\\
{$^1$valeo, $^2$valeo.ai}
\\
{\tt\small \{aimira.baitieva, david.hurych, victor.besnier, olivier.bernard\}@valeo.com}
}
\begin{document}
\maketitle

\newcommand{\DHu}[1]{\textcolor{blue}{[\textbf{DH}: #1]}}
\newcommand{\ABu}[1]{\textcolor{red}{[\textbf{AB}: #1]}}
\newcommand{\VBu}[1]{\textcolor{ForestGreen}{[\textbf{VB}: #1]}}

%%%%%%%%% ABSTRACT
\begin{abstract}

Automating visual inspection in industrial production lines is essential for increasing product quality across various industries. Anomaly detection (AD) methods serve as robust tools for this purpose. However, existing public datasets primarily consist of images without anomalies, limiting the practical application of AD methods in production settings. To address this challenge, we present (1) the Valeo Anomaly Dataset (VAD), a novel real-world industrial dataset comprising 5000 images, including 2000 instances of challenging real defects across more than 20 subclasses. Acknowledging that traditional AD methods struggle with this dataset, we introduce (2) Segmentation-based Anomaly Detector (SegAD). First, SegAD leverages anomaly maps as well as segmentation maps to compute local statistics. Next, SegAD uses these statistics and an optional supervised classifier score as input features for a Boosted Random Forest (BRF) classifier, yielding the final anomaly score. Our SegAD achieves state-of-the-art performance on both VAD (+2.1\% AUROC) and the VisA dataset (+0.4\% AUROC). The code and the models are publicly available\footnote{\url{https://github.com/abc-125/segad}}.

\end{abstract}

%%%%%%%%% BODY
\section{Introduction}
Within the manufacturing process, industrial visual inspection plays a crucial role in identifying defects in produced components. This operation holds significant importance in minimizing costs by identifying and removing faulty parts early in the production stages and, more importantly, in preventing the dispatch of defective components to customers. Traditionally, this task has relied on human operators; however, the likelihood of overlooking certain defects can be as high as 25\% for specific defect types \cite{human_errors}.

When the inspected product comprises numerous components, its examination may create a bottleneck in the production process, causing delays across the entire line. While conventional computer vision methods applied to this task demonstrate superior speed and lower error rates compared to human operators \cite{cv_for_vis_insp}, their inflexibility and lack of satisfactory accuracy \cite{liu2023deep_iad_survey} limit their effectiveness.

%%%%%%%%% TEASER
\begin{figure*}
    \centering
        \includegraphics[width=17cm]{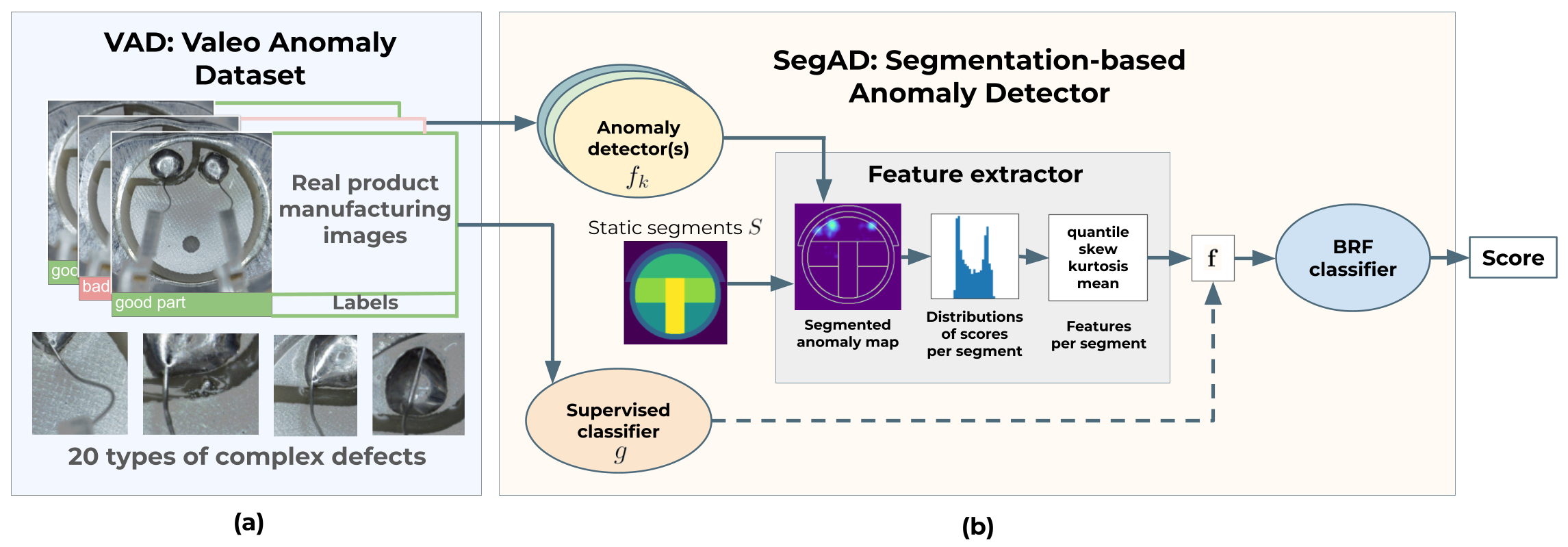}
        \caption{
        Overview of our contributions. \textbf{(a)} VAD, a real-world industrial dataset designed for supervised anomaly detection with complex defects. \textbf{(b)} SegAD, our method that leverages anomaly maps extracted from segmented outputs of one or more anomaly detectors. Higher-level statistical features are computed on these maps, such as skewness, kurtosis, or mean, to generate the local anomaly features. Additionally, our SegAD provides the flexibility to use the output of a supervised classifier score with the local anomaly features, creating input for a final Boosted Random Forest (BRF) classifier that yields the final score.
        }
        \label{fig:teaser}

\end{figure*}        

Industrial deep learning anomaly detection has been an active research field in recent years. Most of anomaly detection methods use only good images for training %(one-class anomaly detection) 
and try to detect deviations from the training data \cite{patchcore, yu2021fastflow, reverse_distillation, efficientad}. Development of the new methods is restrained by publicly available datasets. Recent industrial anomaly detection datasets typically contain approximately one hundred (or even fewer) abnormal images, showcasing defects in the testing set only \cite{zou2022spot, mvtec}. This poses a challenge for supervised anomaly detection methods aiming to utilize both normal and defective parts during training. Supervised models often undergo training with just ten abnormal images from the testing set, resulting in overfitting and reduced sensitivity to previously unseen defects \cite{bgad}. 

In response to the limitations of current datasets, we introduce and publicly release \textbf{VAD} (Valeo Anomaly Dataset),
which contains 1000 bad and 2000 good parts in the training set and 1000 bad and 1000 good parts in the test set, see Fig.~\ref{fig:teaser}(a). Among the defective parts used for testing, 165 contain specific defect types not present in the training data; these parts are explicitly labeled in the released dataset. All images in VAD are captured from an actual production line, showcasing a diverse range of defects, from highly obvious to extremely subtle. This dataset bridges the gap between the academic community and the industry, offering researchers the chance to advance the performance of methods in tackling more intricate real-world challenges.

Current approaches to supervised anomaly detection either yield unsatisfactory results; see Tab.~\ref{tab:hs_sup_benchmark}, or demand pixel-level labels for defective parts, as seen in \cite{bgad, prnad}. Consequently, we introduce a novel method named \textbf{SegAD} (Segmentation-based Anomaly Detector), which is described in Fig.~\ref{fig:teaser}(b).
The employed approach eliminates the need for pixel-level labels, requiring only a flag for each image. Anomaly map scores from each segment are used to delineate the parameters of the distribution, emphasizing relatively low scores in important areas while disregarding higher scores in noisy regions. In this context, the precise position and appearance of the anomaly diminishes in importance as long as it resides within a designated segment.

\noindent{Our main contributions can be summarized as follows:}
\begin{itemize}
    \item We propose a supervised anomaly detection dataset with complex objects and a large variety of defects. As an addition, we establish a one-class anomaly detection benchmark, which is more challenging than most of the current ones, a supervised benchmark with a high number of bad images for training (1000 images) and a supervised benchmark with a low number of bad images for training (100 images). We share the dataset freely with the research community.
    \item We create a novel supervised anomaly detection method called SegAD. This approach performs noticeably better than recent anomaly detectors alone while retaining the ability to detect unknown defects. SegAD reaches SOTA results on VAD as well as on the established VisA dataset~\cite{zou2022spot}.
\end{itemize}

\label{sec:intro}

\section{Related Work}
In this section, we first discuss existing datasets and explain how VAD is different. Next, we give a quick overview of existing anomaly detection methods.
We show in Table \ref{tab:datasets} a global comparison between our VAD and existing datasets.

%------------------------------------------------------------------------
% \begin{figure*}[ht]
\begin{table*}[ht]
\centering
\caption{Comparison of different anomaly detection datasets. 
Tr. good and Tr. bad show the average number of images per class for training. tex = \textit{texture}, obj = \textit{object}, Log. def. = \textit{Logical defects}, Pix. labels = \textit{pixel-level labels} and wl = \textit{weakly labelled} (ellipse around the defect).
}
\label{tab:datasets}
\begin{tabularx}{\textwidth}{lXXXXccccc} 
Dataset & Year & Classes & Images & Types & Defects & Tr. good & Tr. bad & Log. def. & Pix. labels\\ 
\midrule
DAGM \cite{dagm2007}        & 2007 & 10 & 11500 & tex      & synthetic  & 500  & 75   & no  & wl   \\ 
KSSD \cite{kolektorssd2019} & 2019 & 1  & 399   & tex      & real-world & 115  & 17   & no  & yes  \\ 
MVTec AD \cite{mvtec}       & 2019 & 15 & 5354  & tex, obj & simulated  & 242  & 0    & no  & yes  \\ 
KSSD2 \cite{kol2}           & 2021 & 1  & 3335  & tex      & real-world & 2085 & 246  & no  & yes  \\ 
BTAD \cite{btad}            & 2021 & 3  & 2830  & tex, obj & real-world & 600  & 0    & no  & yes  \\ 
LOCO AD \cite{mvtecloco}    & 2022 & 5  & 3644  & obj      & simulated  & 354  & 0    & yes & yes  \\ 
VisA \cite{zou2022spot}     & 2022 & 12 & 10821 & obj      & simulated  & 720  & 0    & no  & yes  \\ 
VAD (ours)                  & 2023 & 1  & 5000  & obj      & real-world & 2000 & 1000 & yes & no   \\ 
\end{tabularx}
\end{table*}
\subsection{Datasets}

In 2019, the MVTec AD dataset \cite{mvtec} was introduced. This dataset significantly advanced anomaly detection methods by providing images with authentic defects and realistic objects, in contrast to earlier datasets predominantly comprising textures and simple defects \cite{dagm2007, kolektorssd2019, kol2}. MVTec LOCO AD \cite{mvtecloco} highlighted the challenge of logical defects, such as missing or misplaced parts, necessitating anomaly detection methods capable of capturing the global context of the image. The more recent \textbf{VisA} dataset \cite{zou2022spot} expands on this by incorporating diverse objects with complex structures, multiple instances, and variations in scale.

Existing datasets often contain simulated defects~\cite{mvtecloco, zou2022spot, mvtec}, creating a domain gap between research and practical applications. This complicates the deployment of anomaly detection methods in real-world settings. To promote supervised anomaly detection, we propose the Valeo Anomaly Dataset (VAD), a real-world industrial dataset with various defects, including logical ones. 

Another notable issue stems from the saturation of solved datasets, where other methods are influenced more by dataset-specific design choices than general applicability.
Compared to this, our dataset introduces a broader range of challenging and diverse anomalies and high intra-class variability of good images.

%------------------------------------------------------------------------
\subsection{Methods}
To define the terms we are working with, “one-class” methods are commonly called unsupervised in the context of anomaly detection \cite{yu2021fastflow, pretrained_feature_extractors, liu2023deep_iad_survey}. We use “one-class” as a more accurate name since these methods use labeled images for training. It is important to distinguish such methods from ones that work with unlabelled data, as in \cite{mcintoshInReaCh}. Models that use bad and good images for training are usually called “supervised”, which is the name we also adopt.

\paragraph{One-class Anomaly Detection.} 
\textbf{PatchCore} \cite{patchcore} relies on a pretrained feature extractor model to extract features from the training set into a memory bank and reduce the size of a memory bank using coreset subsampling. Features extracted from the new input are compared to their nearest neighbors in the memory bank. \textbf{FastFlow} \cite{yu2021fastflow} uses a similar feature extractor model and maps extracted features to the Gaussian distribution using normalizing flow \cite{norm_flows}. \textbf{RD4AD} \cite{reverse_distillation} utilizes a teacher-student architecture, which combines features from several layers of feature extractor to eliminate redundant ones. \textbf{EfficientAD} \cite{efficientad} introduces many innovations, including using an autoencoder to detect logical defects as an addition to a classic teacher-student architecture and their own pretrained feature extractor, which imitates the behavior of a bigger model with a noticeably lesser inference time.

%\paragraph{Supervised Anomaly Detection.}
\textbf{Supervised Anomaly Detection.}
In theory, leveraging defective parts for training is advantageous for refining class boundaries \cite{deepsad}. However, practical challenges arise, such as the small size of anomalies and the impossibility of collecting all potential defects \cite{prnad}.
Several recent anomaly detection methods use both good and bad images \cite{pang2021explainable, bgad, prnad}, but some of them overfit to seen defects, and others require pixel masks for defects to calculate loss or to generate new defects, which can be problematic in a real-world case. Real-world datasets (as VAD) might contain logical defects that cannot be generated by existing defects generation strategies, which usually cut and paste existing defects with various modifications \cite{bgad, prnad}, which is not going to improve results for a wrong wire shape as or other similar types of defects.

Very few supervised anomaly detection benchmarks are available; the most popular is Supervised Anomaly Detection on MVTec AD. 
\textbf{DevNet} \cite{pang2021explainable} is one of the first supervised anomaly detection models that tried to solve industrial MVTec AD dataset, compared to earlier methods which were mostly addressing Out-of-Distribution problem \cite{deepsad, ood_an_det} using non-industrial datasets such as MNIST \cite{lecun2010mnist} or CIFAR-10 \cite{cifar}. \textbf{DRA} \cite{ding2022catching} uses several heads to learn both seen and pseudo anomalies, as well as normal examples, which should reduce overfitting on seen anomalies. These methods perform well on MVTec AD but might fail on more complex problems, as shown in Tab.~\ref{tab:hs_sup_benchmark} and Tab.~\ref{tab:visa}. Compared to modern supervised anomaly detection methods, SegAD (ours) performs better than current SOTA methods, even on complex problems without pixel-level labels, extensive augmentation, and long training.

\label{sec:related_work}

\section{Valeo Anomaly Dataset (VAD)}
\begin{figure}[htp]
    \centering
    \includegraphics[width=8cm]{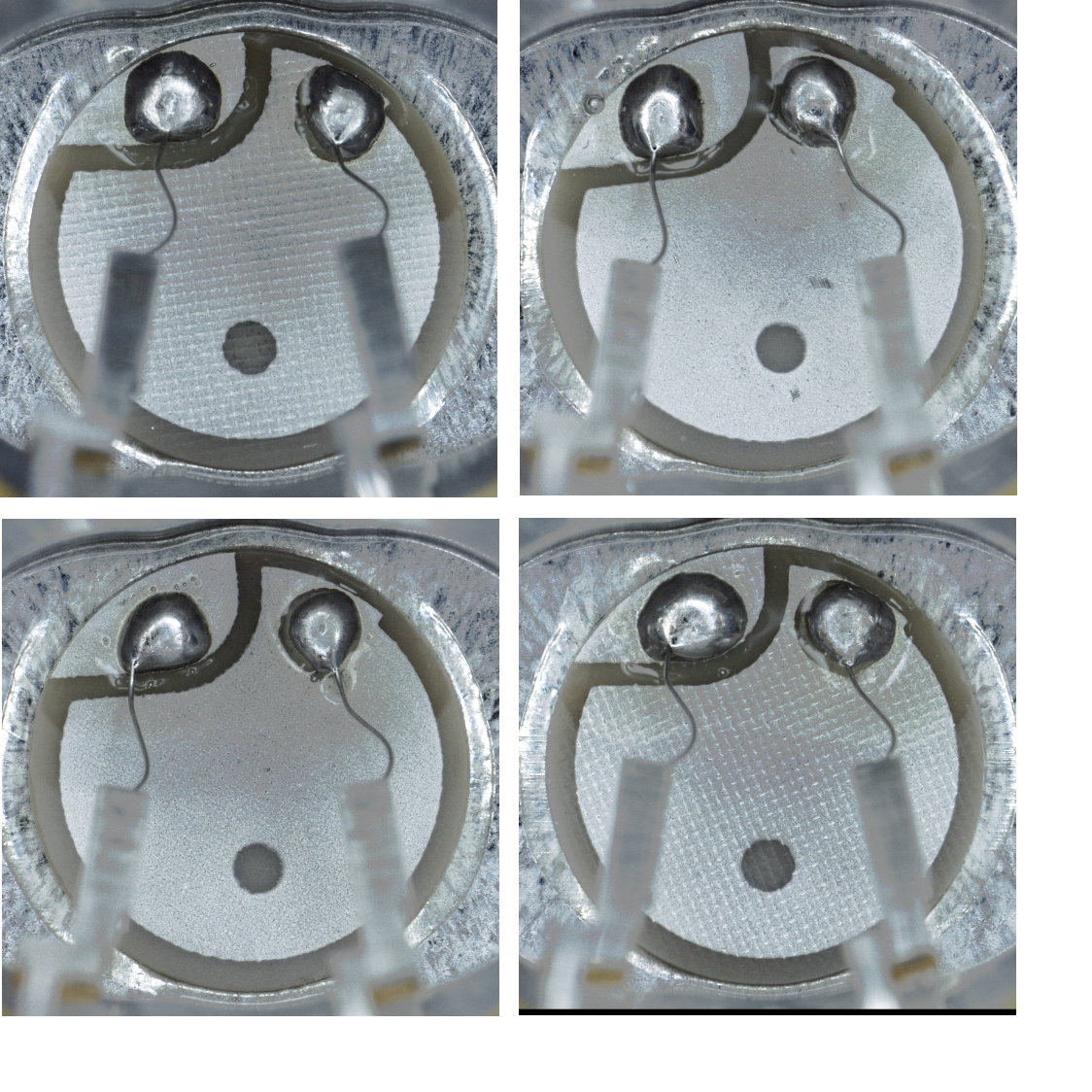}
    \caption{Good parts. Small scratches on the piezo are allowed, as well as wire being closer to the side of the soldering. (VAD)}
    \label{fig:good_parts}
\end{figure}

\begin{figure}[htp]
    \centering
    \includegraphics[width=8cm]{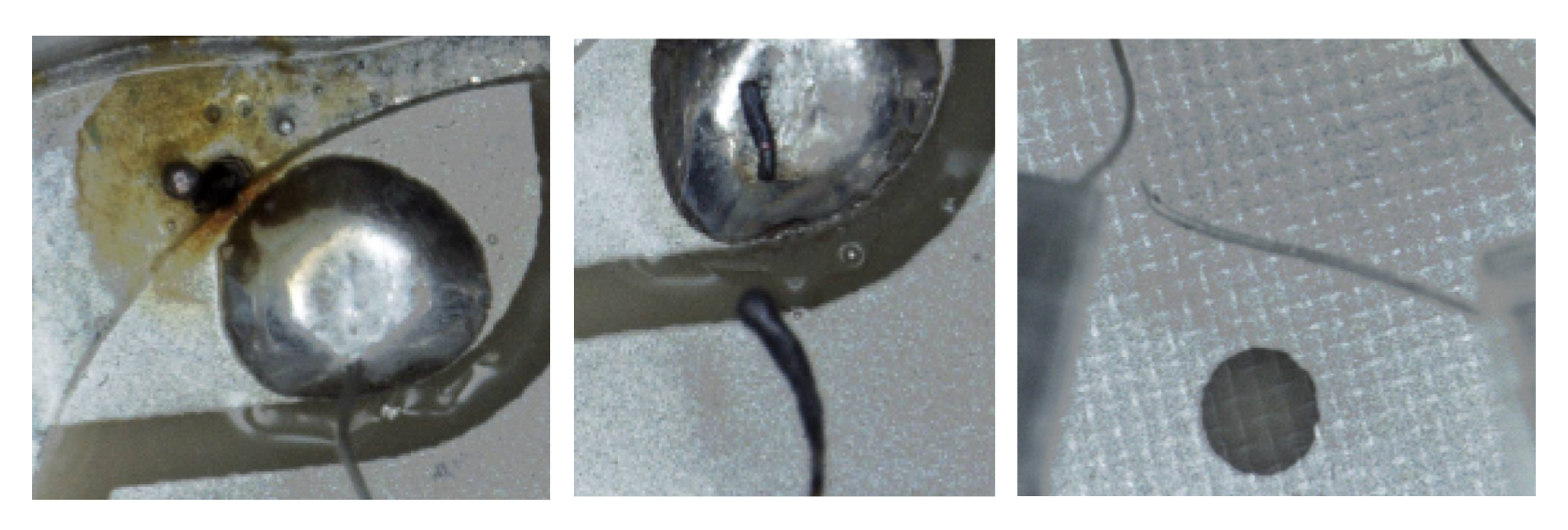}
    \caption{Unseen defects for left to right: burned solder dot, burned wire, extra unnecessary wire. (VAD)}
    \label{fig:unseen_defects}
\end{figure}

VAD consists of one class with predefined training and testing sets. The training set contains 1000 bad and 2000 good images, and the testing set contains 1000 bad, 165 of them are unseen bad, and 1000 good images. Unseen bad images in the test dataset refer to several rare defect types that are not present in the training data, several examples of which are shown in Fig.~\ref{fig:unseen_defects}. Having such images is important to avoid turning this anomaly detection problem into a supervised classification problem.
Some images might have a thin black border at the bottom due to how they were filmed, an example can be seen in Fig.~\ref{fig:good_parts}. Defects might occur in the whole area of the image. Image-level annotation is provided, but there is no pixel-level annotation due to the complexity of defects and the fact there is no exact position for missing or misplaced components. All images are $512\times 512$ pixels in PNG format. Examples of all types of defects can be found in Appendix C.

\begin{figure}[htp]
    \centering
    \includegraphics[width=8cm]{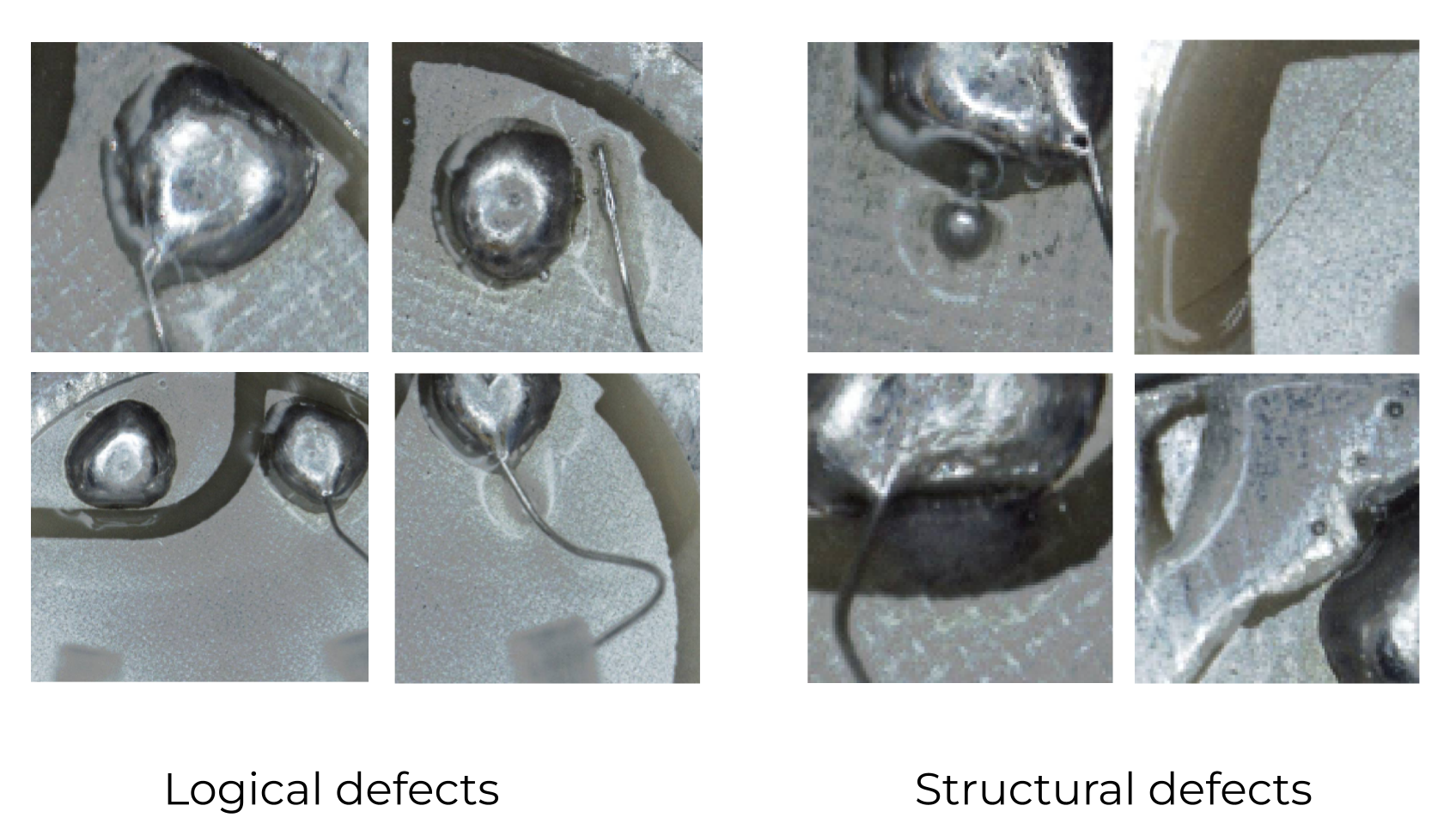}
    \caption{Variety of defects: logical, from the top left: wire on the side, wire out of solder dot, missing wire, bad wire shape (bent too much). Structural: soldering paste pollution, crack on a piezo, burned area under the solder dot, broken piezo. (VAD)}
    \label{fig:defects}
\end{figure}

Good parts consist of two wires connected to two pins on one side and two soldering dots on the other, placed on a piezo. Piezo means piezoelectric element, a big round area under the other components.
Wires should have the right shape, not too straight, too long, or bent too much. Wires should connect the soldering dots as close to their centers as possible. Some amount of deviation is allowed, but if the wire is too close to the side of a solder dot, it is a defect. Soldering should be well-rounded, not too small, and placed in the correct location, not overlapping with the piezo contours. The wire should be properly connected to the pin without any pieces of it visible on the side of the pin. The piezo can have different textures. Any kind of pollution is considered to be an anomaly, which makes the part defective. 

Logical defects result from components being misplaced or misshapen rather than being damaged, e.g. bad wire shape, bad soldering shape, wrong wire position, wrong soldering position, and missing parts. Structural defects include various cracks and broken parts of the piezo, as well as a variety of burns and pollution. For both types of defects, see Fig. \ref{fig:defects}. The importance of logical defects is often overlooked in the existing datasets (excluding MVTec LOCO AD) because they are hard to imitate if the dataset is filmed in the lab, but also hard to detect. VAD provides the research community the opportunity to work with real logical constraints of a complex object.

%------------------------------------------------------------------------
\subsection{Benchmarks}
\label{subsection:VAD_benchmarks}
We establish three benchmarks for the VAD dataset. The first is \textbf{one-class anomaly detection}. As shown in Tab~\ref{tab:one_class_benchmark}, even SOTA methods struggle to perform well on our dataset compared to other popular datasets, which leaves an opening for further research. For this benchmark, no bad images are used for training. The second benchmark is \textbf{high-shot supervised anomaly detection}, which uses all 1000 bad images from the training set. The last benchmark is \textbf{low-shot supervised anomaly detection}. Only 100 bad images from the training set are used. These images are randomly selected using 5 seeds, with the average result over all seeds being expected. The same 2000 good parts are used for the training in all benchmarks.

\label{sec:dataset}

\section{Method}
%------------------------------------------------------------------------
%\subsection{Description}
SegAD works with image data at its input, and after inference, it is expected to output a score that describes a probability of anomaly appearance. Let $\mathcal{I} = \bigr \llbracket 0, \dots, 255\bigr \rrbracket^{W,H,3}$ be a space of all images of width $W$, height $H$ and with 3 color channels. The $i-$th image from a dataset is marked as $I^{(i)}\in \mathcal{I}$. We denote $\mathbf{x}$ to be a 2D pixel coordinate vector. SegAD inference consists of three consecutive stages in a pipeline. First is anomaly map(s) calculation with optional supervised classifier score calculation. Next is the calculation of simple statistics from anomaly maps over individual segments. Lastly, the obtained statistics, and optionally also the classifier score, are used as input features for a BRF classifier~\cite{boosted_rand_forest} that delivers the final result.

Let us denote $K$ as the number of used pixel-wise anomaly detectors and mark them as functions $f_k(I): \mathcal{I} \rightarrow \mathbb{R}^{W,H}, k\in \{1, \dots, K\}$. A supervised classifier is a function $g(I):\mathbb{R}^{W,H} \rightarrow \mathbb{R}$. Next, we introduce a set of $L$ mutually exclusive segments $S=\{ s_1, \dots, s_L\}$, where $s_l \in \bigr \llbracket 0, 1\bigr \rrbracket^{W,H}$.

After the calculation of anomaly maps for all $f_k(I)$, the masks $s_l$ are used for segment-wise extraction of results, followed by the calculation of several statistics.
The final feature vector $\mathbf{f}=\left[g(I), [\mathbf{q}^L]_{k=1}^{K}, [\mathbf{z}^L]_{k=1}^{K}, [\mathbf{c}^L]_{k=1}^{K}, [\mathbf{m}^L]_{k=1}^{K} \right]$, extracted from one image, is the concatenation of 4 vectors containing different statistics. $\mathbf{q}^L_k$ stands for a vector of $L$ numbers where each stands for a $99.5\%$ quantile from anomaly map $f_k(I)$ over pixels where $s_l(\mathbf{x}) = 1$. Similarly, for every anomaly map $f_k(I)$ we calculate a vector for skew $\mathbf{z}^L_k$, kurtosis $\mathbf{c}^L_k$ and mean $\mathbf{m}^L_k$. The length of the feature vector $\mathbf{f}$ equals to $K*L*4+1$. 

For notation simplicity, we keep the supervised classifier score $g(I)$ always present in $\mathbf{f}$, but it may be omitted, which is clearly defined in every experiment setup. Feature vectors $\mathbf{f}^{(i)}$ are used as inputs for training the final classifier~\cite{boosted_rand_forest}, which, at inference time, delivers the final result. 

%------------------------------------------------------------------------
\subsection{Training}
\label{subsection:training}
Training SegAD requires two separate subsets from the training dataset. The first subset is used to train base models $f_k$, the second subset is used to train the BRF classifier. To create these subsets from one training dataset, we split available good images into two parts. For VAD, we report average results over 5 different splits to show more reliable results.

Each model ($f_k, g$, BRF) is trained independently. Training can be separated into two stages. In the first stage, anomaly detector(s) $f_k$ are trained using the same data. In the same stage, a segmentation map $S$ must be defined. As will be elaborated in more detail in Sec. \ref{sec:experiments}, the segmentation map can be static or produced by a segmentation model, depending on the dataset. A supervised classifier $g$ is trained separately on 80\% of the full training set (1600 good and 800 bad parts for VAD) to ensure that the training sets for this model and SegAD are not fully overlapping. The next stage is to train a final BRF classifier in SegAD, which requires both good and bad images, bad images can be replaced with artificial defects in some cases, as shown in Tab. \ref{tab:one_class_benchmark}. BRF is trained with the output of the models from the previous stage. 

An important thing to note is that SegAD is very fast to train as long as you already have other models available. Extraction of features and training the final classifier takes noticeably less time than training anomaly detectors. The same is true for the inference; SegAD adds a negligible time to processing the image while improving results significantly.

\subsection{Setups for SegAD} 
\label{subsection:setups}
SegAD can use any anomaly detector(s) $f_k$ as long as it produces an anomaly map showing pixel-level anomaly scores, although, in this paper, we use only one-class anomaly detectors because they require less data to train. Any supervised classifier $g$ that returns a classification score can be used. In experiments, we use several setups, which will be described here in detail. SegAD is denoted as Ours. Supervised classifier $g$ is not used unless stated otherwise.

\textbf{Anomaly Detector + Ours} means $K=1$, $f_1 =$ “Anomaly Detector name”. This setup allows us to show improvement for one anomaly detection model in particular and demonstrate that this improvement can be achieved for different types of anomaly detectors. 

\textbf{All AD + Ours} denotes that all anomaly detectors from the list $AllAD$ of length $N$ are used to produce anomaly maps. Such a list can be found in the caption of the table. In that case $K=N$, $f_1 = All AD_1, \dots, f_N = All AD_N$. Anomaly detectors might use an ensemble of feature extractors, similarly to \cite{patchcore}, which makes them slower but can improve the result. In our case, anomaly detectors themselves can be successfully assembled in a supervised way. 

\textbf{All AD + Supervised Classifier + Ours} denotes a setup in which $g = $ “Supervised Classifier name”, and the rest is the same as in the previous setup. A supervised classifier allows us to detect seen defects, which can be invisible to a one-class classifier, similarly to \cite{ind_framework}. Such a setup shows the best result for academic purposes, but the speed of inference can be unsatisfying for real-world applications. We expect that, in that case, the number and selection of anomaly detectors can be optimized to fit the problem as well as the time constraints.

\textbf{Anomaly Detector + Supervised Classifier + Ours} mean a setup in which $g = $ “Supervised Classifier name”, $K=1$, $f_1 = $ “Anomaly Detector name”. It shows an example of an optimization mentioned in the previous setup. Only one anomaly detector gives a higher inference speed compared to the ensemble of anomaly detectors while giving a satisfying result.
\label{sec:method}

\section{Experiments}
\textbf{Evaluation metrics.} We compare results using the following metrics: (1) Area Under the Receiver Operating Characteristic Curve (AUROC) \cite{auroc}, (2) False Positive Rate (FPR) at 95\% True Positive Rate (TPR) denoted as FPR@95TPR.
Classification AUROC, which describes the performance on the image level for multiple thresholds, denoted by Cl. AUROC. 
FPR@95TPR shows the percentage of misclassified good parts (false positive) at 95\% of bad parts classified correctly (true positive). We report mean results $\pm$ standard deviation.

\textbf{Implementation and evaluation details.} 
SegAD uses the output of one-class detectors PatchCore \cite{patchcore}, FastFlow \cite{yu2021fastflow}, Reverse Distillation \cite{reverse_distillation} (referred to as RD4AD), EfficientAD \cite{efficientad} either alone or several anomaly detectors at once. SegAD results are compared to the results of these models alone, as well as supervised anomaly detectors DevNet \cite{pang2021explainable} and DRA \cite{ding2022catching}. We also compare our results to the supervised classifier Wide ResNet50 \cite{wrn_clf} (denoted as WRN). The final classifier BRF was trained with the same parameters for all datasets using XGBoost library \cite{xgboost}. More details on the implementation can be found in Appendixes A and B. 

Images are resized to W = H = 256 pixels to make the comparison less biased. The DRA model originally uses a resolution $448\times 448$, which shows better results on VAD than $256\times 256$, 92.8 Cl. AUROC, which is still lower than the top results. The higher resolution also improves results for some of the other models, so we do not use it. $256\times 256$ is sufficient for defects to remain visible, but it also allows to have a reasonable speed of inference, especially for anomaly detection models.

For VAD, results averaged for 5 seeds and training runs are reported. These seeds are used for training and splitting the training set for SegAD into two sets, as described in Subsection~\ref{subsection:training}. 2000 good parts from the train set were split in half: 1000 for training the base model(s) and 1000 for training SegAD. 0, 1000, or 100 bad parts (depending on the benchmark) were also used to train SegAD. Details on the VisA benchmark can be found in Subsection \ref{subsection:results_visa}
No augmentations were used for training, and existing augmentations were removed from used methods. 
Augmentations tend to be suitable only for specific tasks (as rotation in DRA or DevNet, removing it improved results by these models on the VAD drastically), and using augmentations makes it hard to compare models themselves.

We include three benchmarks for the new \textbf{VAD} and one additional for the \textbf{VisA} dataset \cite{zou2022spot}. We compare results for VAD for one-class, high-shot, and low-shot supervised benchmarks, as described in Subsection \ref{subsection:VAD_benchmarks}. The benchmark for the VisA dataset is explained in \ref{subsection:results_visa}.
\begin{figure}[htp]
    \centering
    \includegraphics[width=5cm]{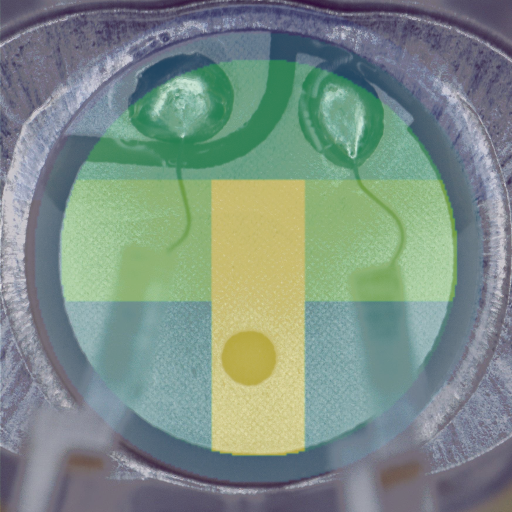}
    \caption{Static segmentation map for objects in VAD dataset, overlayed over the image. The image is separated into $L=7$ segments: background, outer half-circle, piezo border, solder dots area, wires area, pins area, and piezo in the middle.
    \label{fig:seg_map_VAD}}
\end{figure}
The segmentation maps for VAD is the same static image for every input image from the dataset. The segmentation map can be seen in Fig. \ref{fig:seg_map_VAD}. Due to the fact that the object in the VAD is always centered with the constant size and component placement, this is the simplest choice, which may be sub-optimal, but serves the purpose and does not require any manual or model-inferred annotation of the data. It was created to separate the anomaly maps into several meaningful parts, such as piezo border, wire areas, soldering area, etc., which can contain different defects and different levels of falsely high scores on the anomaly map. 
Another source of segmentation maps can be a specially trained segmentation model or a zero-shot segmentation model.

%------------------------------------------------------------------------
\subsection{VAD, one-class benchmark}
This benchmark allows only good images for training, but SegAD requires bad images as well, so we try to replace them with artificially generated defects. Because SegAD works with anomaly maps, generating realistic defects is unnecessary. For this reason, we have applied several simple augmentations to the 1000 good images available for training SegAD. These augmentations include Gaussian blur or a randomly placed, randomly sized thin rectangle of a random shade of gray. They were applied to randomly selected segments or random parts of the segments; segments are defined with the segmentation map. Code for generating defects can be found in the SegAD GitHub repository.

\begin{table}[htp]
\centering

\begin{tabularx}{\columnwidth}{Xcc} 
    Method & Cl. AUROC $\uparrow$ & FPR@95TPR $\downarrow$\\ 
    \midrule
    \multicolumn{3}{c}{\textit{One-class Anomaly Detection}} \\
    \midrule
    FastFlow & 79.1\small{$\pm$1.7} & 76.2\small{$\pm$1.9}\\
    RD4AD & 84.5\small{$\pm$0.2} & 65.8\small{$\pm$2}\\
    PatchCore & 88.3\small{$\pm$0.3} & 60.4\small{$\pm$2.8}\\
    EfficientAD & 89.1\small{$\pm$0.8} & \textbf{44.5}\small{$\pm$2.2}\\
    \midrule
    \multicolumn{3}{c}{\textit{SegAD (Ours)}} \\
    \midrule
    FastFlow + Ours & 83.3\small{$\pm$1.3} & 71.5\small{$\pm$5}\\
    RD4AD + Ours & 87.5\small{$\pm$1.2} & 61.0\small{$\pm$7.9}\\
    PatchCore + Ours & 89.7\small{$\pm$0.8} & 58.8\small{$\pm$3.7}\\
    EfficientAD + Ours & 85.6\small{$\pm$1.6} & 100\small{$\pm$0}\\
    All AD + Ours & \textbf{90.4}\small{$\pm$0.5} & 52.4\small{$\pm$4.5}\\
\end{tabularx}

\caption{One-class benchmark (VAD). The best result is marked in bold. All AD means PatchCore, FastFlow, and RD4AD. \label{tab:one_class_benchmark}}
\end{table}

Results in Table~\ref{tab:one_class_benchmark} show that even such a naive approach improves results compared to anomaly detection methods alone. Even greater improvement can be achieved by using several anomaly detectors to produce anomaly maps for SegAD. This strategy fails for EfficientAD + Ours, anomaly maps produced by EfficientAD for the generated defects are too different from anomaly maps for real defects, creating a large discrepancy between training and test distributions for SegAD and causing worse results compared to anomaly detector alone.

%------------------------------------------------------------------------
\subsection{VAD, high-shot supervised benchmark}
\begin{table}[htp]
\centering
\begin{tabularx}{\columnwidth}{Xcc} 
Method & Cl. AUROC $\uparrow$ & FPR@95TPR $\downarrow$\\ 
\midrule
\multicolumn{3}{c}{\textit{Supervised Anomaly Detection}} \\
\midrule
DevNet & 86.9\small{$\pm$1.2} & 66.6\small{$\pm$2.7}\\
DRA & 87.4\small{$\pm$0.6} & 60.9\small{$\pm$4.4}\\
\midrule
\multicolumn{3}{c}{\textit{Supervised Classifier}} \\
\midrule
WRN & 95.0\small{$\pm$0.6} & 32.8\small{$\pm$4.3}\\
\midrule
\multicolumn{3}{c}{\textit{SegAD (Ours)}} \\
\midrule
FastFlow + Ours & 86.9\small{$\pm$0.5} & 56.9\small{$\pm$3.9}\\
RD4AD + Ours & 90.2\small{$\pm$0.3} & 46.9\small{$\pm$2.6}\\
PatchCore + Ours & 91.7\small{$\pm$0.3} & 48.0\small{$\pm$2.9}\\
EfficientAD + Ours & 91.4\small{$\pm$0.2} & 36.8\small{$\pm$1}\\
AllAD+WRN+Ours & \textbf{96.5}\small{$\pm$0.3} & \textbf{16.4}\small{$\pm$1.9}\\
\end{tabularx}
\caption{High-shot supervised benchmark (VAD). The best result is marked in bold. All AD means PatchCore, FastFlow, RD4AD, and EfficientAD. Improvement calculated compared to base method results in Tab.~\ref{tab:one_class_benchmark}.
\label{tab:hs_sup_benchmark}}
\end{table}

\begin{figure}[htp]
    \centering
    \includegraphics[width=8cm]{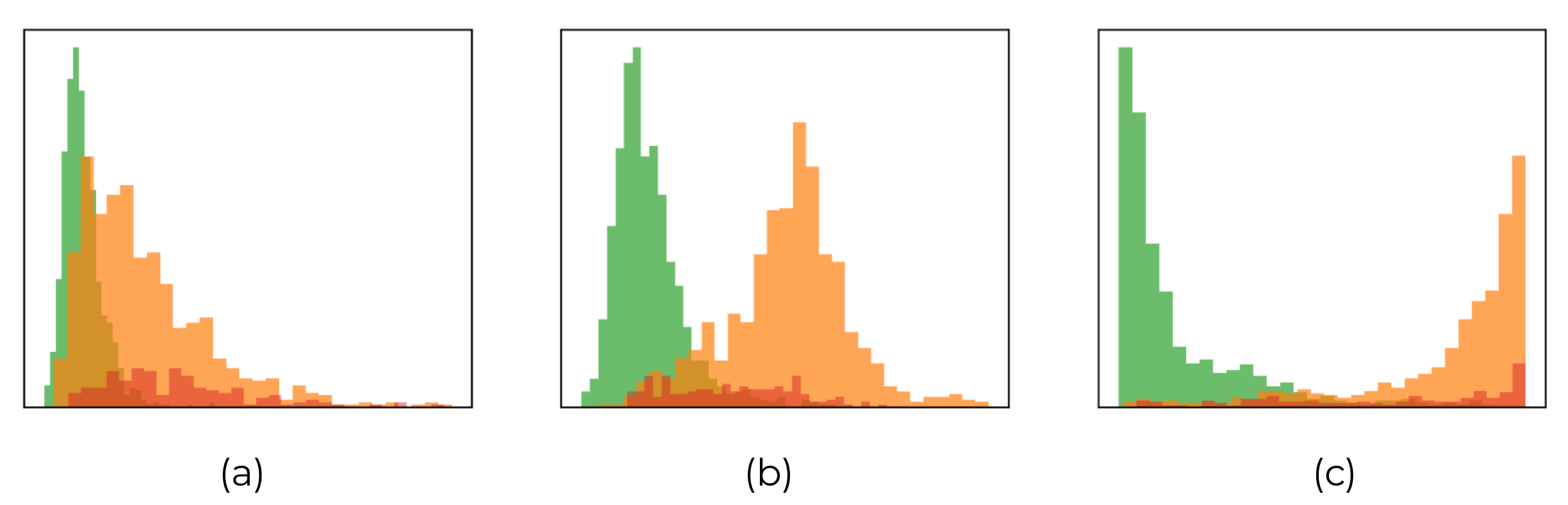}
    \caption{Distributions of scores in the VAD test set. Green color shows good parts, orange bad parts, and red bad parts with unseen defects. (a) shows one-class anomaly detector PatchCore, (b) supervised classifier Wide ResNet, (c) SegAD (PatchCore + WRN + Ours). X axis is score values, Y axis is frequency.}
    \label{fig:histograms}
\end{figure}

The high-shot benchmark allows training supervised classifier $g$, which shows competitive results as can be seen in Tab.~\ref{tab:hs_sup_benchmark} in the row “WRN”, but it won't be able to detect unseen defects. Histograms in Fig. \ref{fig:histograms} visualize this problem. On the left is the one-class anomaly detector, which can detect unseen defects (in red) but shows a bad separation between good and bad parts. In the center, the supervised classifier $g$ shows a better separation of good and bad parts, but unseen defects distribution is shifted to the left because their scores are relatively low and such defects cannot be detected. On the right, SegAD shows a good separation between classes, but also unseen defects have a similar distribution to the seen defects. We still compare the supervised classifier Wide ResNet50 (WRN) with other methods, but the ability to detect unseen defects is crucial in real-world applications.

Combining SegAD, supervised classifier WRN, and all one-class anomaly detection models showed the best results presented in Table~\ref{tab:hs_sup_benchmark}. A more practical setting can be EfficientAD + WRN + Ours, which can achieve Cl. AUROC 96 and FPR@95TPR 19.4 with a much higher inference speed.

%------------------------------------------------------------------------
\subsection{VAD, low-shot supervised benchmark}
This benchmark uses just 100 bad images for training, which can be a relatively low number for a dataset with more than 20 types of defects. However, such a setting is closer to real-world applications, where a limited amount of defective images can be available. In Table~\ref{tab:ls_sup_benchmark}, WRN shows much lower results due to a smaller training dataset, as well as DevNet and DRA. SegAD also performs worse, than with a high-shot benchmark, but it still can visibly improve the performance of other models. We do not use WRN in SegAD for this benchmark, because it showed low results by itself.

\begin{table}[htp]
\centering
\begin{tabularx}{\columnwidth}{Xcc} 
Method & Cl. AUROC $\uparrow$ & FPR@95TPR $\downarrow$\\ 
\midrule
\multicolumn{3}{c}{\textit{Supervised Anomaly Detection}} \\
\midrule
DevNet & 73.1\small{$\pm$2.3} & 88.3\small{$\pm$2.3}\\
DRA & 78.9\small{$\pm$1.8} & 74.6\small{$\pm$1.9}\\
\midrule
\multicolumn{3}{c}{\textit{Supervised Classifier}} \\
\midrule
WRN & 78.7\small{$\pm$1.8} & 77.9\small{$\pm$4.4}\\
\midrule
\multicolumn{3}{c}{\textit{SegAD (ours)}} \\
\midrule
FastFlow + Ours & 84.3\small{$\pm$1.4} & 64.5\small{$\pm$2.5}\\
RD4AD + Ours & 88.6\small{$\pm$0.2} & 51.1\small{$\pm$0.9}\\
PatchCore + Ours & 90.8\small{$\pm$0.5} & 52.7\small{$\pm$4.3}\\
EfficientAD + Ours & 90.6\small{$\pm$0.7} & 40.1\small{$\pm$2.4}\\
All AD + Ours & \textbf{92.7}\small{$\pm$0.5} & \textbf{36.8}\small{$\pm$3.1}\\
\end{tabularx}
\caption{Low-shot supervised benchmark (VAD). The best result is marked in bold. All AD means PatchCore, FastFlow, RD4AD, and EfficientAD.
\label{tab:ls_sup_benchmark}}
\end{table}

%------------------------------------------------------------------------
\subsection{Results on VisA dataset}
\label{subsection:results_visa}
\begin{figure}[htp]
    \centering
    \includegraphics[width=8cm]{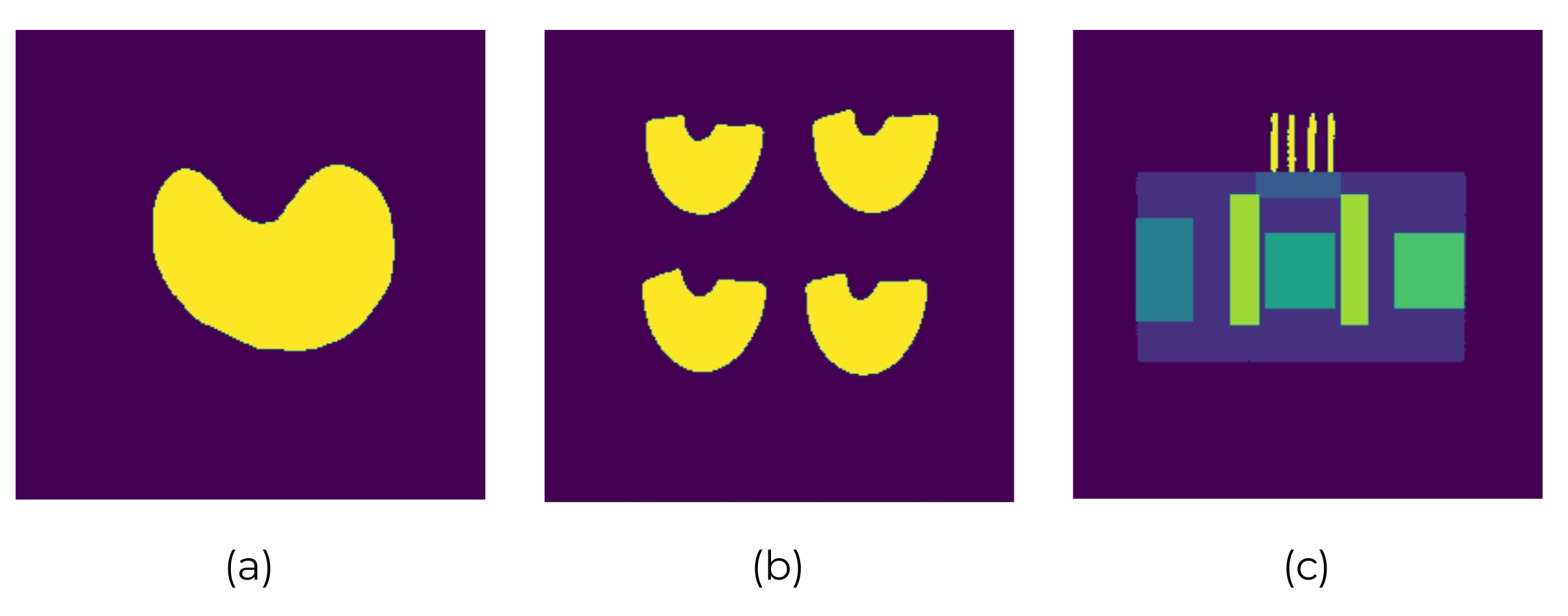}
    \caption{VisA. Examples of segmentation maps created using SAM for (a) cashew, (b) macaroni1, (c) pcb2.}
    \label{fig:seg_masks_visa}
\end{figure}

VisA contains fewer good images per class for training than VAD (500-1000 in VisA, compared to 2000 in VAD). The training set consists of good images only. For this reason, we use a different strategy to split the training dataset to make sure we have enough images to train the base anomaly detection model with a competitive result. For RD4AD, the training set was divided into 90\% of images to train the RD4AD itself and 10\% to train SegAD. EfficientAD uses a validation set (10\% of good images from the train set) to normalize anomaly maps. In our case, the same images were used to train SegAD. 
The segmentation maps for this dataset were created with the Segment Anything Model (SAM) \cite{kirillov2023segany}. For most classes, it is a binary mask for the object and background, and for pcb1-4, static components were mapped on top of it; see Fig. \ref{fig:seg_masks_visa}.

\begin{table}[htp]
\centering
\begin{tabularx}{\columnwidth}{Xcc} 
Method & Cl. AUROC $\uparrow$ & FPR@95TPR $\downarrow$\\ 
\midrule
\multicolumn{3}{c}{\textit{One-class Anomaly Detection}} \\
\midrule
RD4AD & 94.7 & 24.5\\
EfficientAD & 97.9 & 8.3\\
\midrule
\multicolumn{3}{c}{\textit{Supervised Anomaly Detection}} \\
\midrule
DRA & 88.9 & 42.4\\
DevNet & 89.3 & 44.0\\
\midrule
\multicolumn{3}{c}{\textit{SegAD (ours)}} \\
\midrule
RD4AD + Ours & 95.3 & 19.3\\
EfficientAD + Ours & 98.3 & 8.1\\
All AD + Ours & \textbf{98.4} & \textbf{7.8}\\
\end{tabularx}
\caption{Results on VisA dataset, supervised benchmark. The best result is marked in bold. All AD means RD4AD and EfficientAD.
\label{tab:visa}}
\end{table}

With VisA, we create a new, challenging, supervised benchmark on a public dataset with 12 different classes. In a similar way to the supervised benchmark on MVTec AD \cite{pang2019deep}, we move 10 randomly selected bad images from the test set to the train set, reporting results over 10 different seeds. In Table~\ref{tab:visa}, this benchmark shows how SegAD performs on a different data from VAD and creates a new possibility to compete on a difficult supervised anomaly detection problem. Detailed results per class can be found in Appendix D. 

%------------------------------------------------------------------------
\subsection{Ablation study}

We perform an ablation study on the VAD high-shot supervised benchmark. We use the average result of the PatchCore, FastFlow, RD4AD, and EfficientAD as the baseline in the Table~\ref{tab:abl_segad}, denoted as “An.Det.”. These anomaly detectors use the maximum value from the anomaly map to calculate the score. On the contrary, we compute our proposed features $\mathbf{f}$, which are described in Section~\ref{sec:method}, and use BRF to yield a final score. This shows an improvement of 1.3 Cl. AUROC in the row “One Seg.”. In addition, we evaluate the segmentation map's impact by computing the anomaly map's maximum value per segment.
Doing this gives an improvement in 2.1 Cl. AUROC, which can be seen in the row “Max.”. This shows how both features and segmentation maps are important for SegAD. We also show the strength of BRF by comparing it against other algorithms. We compare BRF with BT and RF using a segmentation map and features. The results show that the BRF is better than these methods by 1.7 and 0.3 Cl. AUROC, respectively.

\begin{table}[htp]
\centering
\begin{tabularx}{\columnwidth}{c|ccc|c} 
Description & Feat. $\mathbf{f}$ & Seg. $S$ & BRF & Cl. AUROC $\uparrow$ \\ 
\midrule
An. Det. & & & & 87.3\\
\midrule
One Seg. & \Checkmark &  & \Checkmark & 88.6\\
Max. & & \Checkmark & \Checkmark & 89.4\\
BT & \Checkmark & \Checkmark & & 88.4\\
RF & \Checkmark & \Checkmark & & 89.8\\
\midrule
SegAD & \Checkmark & \Checkmark & \Checkmark & 90.1\\
\end{tabularx}
\caption{Ablation study on the methodical differences between our method (SegAD) and anomaly detection methods. An.Det. = \textit{Anomaly Detectors}. One Seg. = \textit{One Segment}. Max. means using maximum value. BT = \textit{Boosted Tree}. RF = \textit{Random Forest}. BRF = \textit{Boosted Random Forest}. SegAD denotes the average result from PatchCore + Ours, FastFlow + Ours, RD4AD + Ours, EfficientAD + Ours.
\label{tab:abl_segad}}
\end{table}

\paragraph{Limitations:} SegAD (ours) requires to have segmentation maps. Static maps can be used for objects like the product in VAD. Obtaining segmentation maps for unaligned objects can be more difficult, yet SAM~\cite{kirillov2023segany} or a specially trained segmentation model may be a solution. The performance improvement of SegAD depends on the complexity and structure of the objects. For less complex objects in VisA, improvement is lower than for VAD, as can be seen in Tab.~\ref{tab:hs_sup_benchmark} and Tab.~\ref{tab:visa}. Another possible limitation can be the difference in the anomaly maps between training and test data. Due to the nature of anomaly detection, this problem might be impossible to mitigate, because our task is to detect such differences.

\label{sec:experiments}

\section{Conclusion}
This study introduces VAD, a brand new supervised anomaly detection dataset derived from real production, offering challenging benchmarks for the research community to address real-world defect detection. Additionally, we propose SegAD, an innovative supervised anomaly detection method that achieves state-of-the-art performance on VAD as well as on a supervised benchmark on standard VisA dataset. Experimental results reveal the limitations of existing supervised anomaly detection methods in handling complex problems, while the integration of SegAD with top-performing one-class anomaly detection models further enhances their results.

\label{sec:conclusion}

%%%%%%%%% REFERENCES
{\small
\bibliographystyle{ieee_fullname}
\bibliography{egbib}
}

%%%%%%%%% Appendices
\onecolumn
\section{Appendices}
\appendix

%--------------------------------------------------------------------------------
\section{Implementation Details for SegAD} 
\label{appendix:impl}
This section describes implementation details for the final classifier (Boosted Random Forest) in SegAD and shows several examples of generating defects for the VAD one-class benchmark. Models were trained with NVIDIA A2000. More details, the code, segmentation maps, and trained models can be found in SegAD GitHub repository\footnote{\url{https://github.com/abc-125/segad}}.

\subsection{Final Classifier}
We use the same Boosted Random Forest (BRF) for all experiments with SegAD, except for the Ablation Study, which also uses Random Forest (RF) and Boosted Tree (BT). Implementation by XGBoost \cite{xgboost} is used for all three of them. \textbf{BRF} in all experiments uses 10 estimators and 200 parallel trees with a learning rate of 0.3 and \emph{binary:logitraw} objective. Maximum depth is set to 5 to avoid overfitting; for the same reason, we set the subsample ratio of columns when constructing each tree (\emph{colsample\_bytree}) to 0.6, the same as the subsample ratio per node (\emph{colsample\_bynode}) and \emph{subsample}. We also use the L1 regularization \cite{lasso_l1} with a value of 1. For \textbf{RF}, the learning rate is 1.0, the number of estimators is 1, and the number of trees is 2000.
For \textbf{BT}, the number of estimators is 2000, the number of trees is 1, \emph{colsample\_bytree} is default.

\subsection{Generating Defects}
\begin{figure}[htp]
    \centering
    \includegraphics[width=8cm]{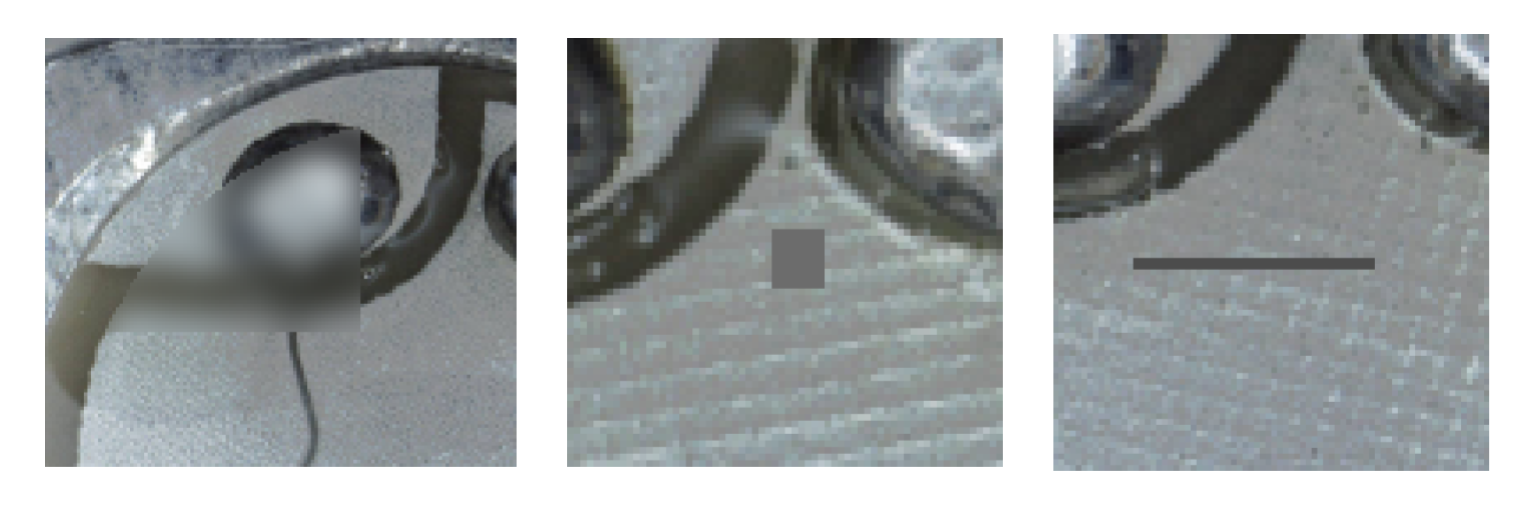}
    \caption{Generated defects, from the left: part of the segment is blurred, small rectangle, thin rectangle. (VAD)}
    \label{fig:bad_generated}
\end{figure}

Generating defects for the one-class benchmark is not optimized and is used just as a demonstration that SegAD can work with one-class tasks as well. In our case, it was sufficient to generate disturbances in the image, which caused anomaly detectors $f_k$ to show higher anomaly scores in their output, which was sufficient to train SegAD. The code to generate defects can be found in the SegAD GitHub repository. Several examples of generated defects can be seen in Fig.~\ref{fig:bad_generated}. Advanced strategies for generating defects, which are described in \cite{han2022adbench, draem}, can be investigated in future work. 

%--------------------------------------------------------------------------------
\section{Implementation Details for Other Evaluated Methods}
\label{appendix:impl_others}
In this section, we describe implementation details and training parameters for other methods used in this paper. All models use resolution $256\times 256$ and pretrained feature extractor WideResNet-50-2 by TorchVision \cite{paszke2017automatic} unless stated otherwise. No augmentations were used. All existing augmentations were removed to have a more fair comparison and to avoid fitting a specific problem. These removed augmentations include center crop for PatchCore, which improves results as long as defects are close to the center of the image, e.g., for MvTec AD, see EfficientAD paper \cite{efficientad} for more details and random rotation for DRA and DevNet which made their results worse for VAD. Models were trained without early stopping. 

\textbf{Wide ResNet.}
WideResNet-50-2, pretrained weights by TorchVision \cite{paszke2017automatic}. Trained with a batch size of 4, the learning rate is 0.025, uses focal loss \cite{focal_loss}. The optimizer is SGD with a momentum of 0.9.

\textbf{PatchCore.}
Unofficial implementation by Akcay et al.~\cite{anomalib}. The feature extractor uses layers 2 and 3. The coreset sampling ratio is 0.01, and the number of neighbors is 9.

\textbf{Reverse Distillation.}
Unofficial implementation by Akcay et al.~\cite{anomalib}. The feature extractor uses layers 1, 2, and 3. \emph{Beta1} is 0.5, \emph{beta2} is 0.999. Anomaly maps are combined through addition. Trained for 200 epochs.

\textbf{FastFlow.}
Unofficial implementation by Akcay et al.~\cite{anomalib}. Trained for 200 epochs (similar as in EfficientAD paper \cite{efficientad}), learning rate is $1e-3$, weight decay is $1e-5$.

\textbf{EfficientAD.}
Unofficial implementation by Nelson\footnote{\url{https://github.com/nelson1425/EfficientAD}}, the original code was not published. Trained for 70000 iterations without an early stopping. 10\% of the training dataset was used for normalization. Uses padding in convolutional layers. The model size is medium.

\textbf{DevNet.}
Official implementation by Pang et al. \cite{pang2019deep}. The feature extractor is ResNet18 by TorchVision \cite{paszke2017automatic}. For VAD, the code was modified to use bad images from a separate folder instead of a testing set. For VisA, the code was modified to split
bad images from the test set with predefined seeds, the same as for SegAD (ours). The parameter \emph{n\_anomaly} (number of bad images used for training) was set to 1000 for high-shot, 100 for low-shot VAD, and 10 for VisA. Trained for 50 epochs.

\textbf{DRA.}
Official implementation by Ding et al. \cite{ding2022catching}. The feature extractor is ResNet18 by TorchVision \cite{paszke2017automatic}. The code was modified similarly to DevNet, with the same values for \emph{n\_anomaly}. %Using lower values for \emph{n_anomaly} showed worse results. - need numbers
Trained for 30 epochs.

%--------------------------------------------------------------------------------
\section{Additional details on VAD}
\label{appendix:vad}
This section shows the detailed scheme of the product from VAD and the list of defects with examples. This can be useful for a better understanding of the complexity of the dataset. 

\begin{figure}[htp]
    \centering
    \includegraphics[width=6cm]{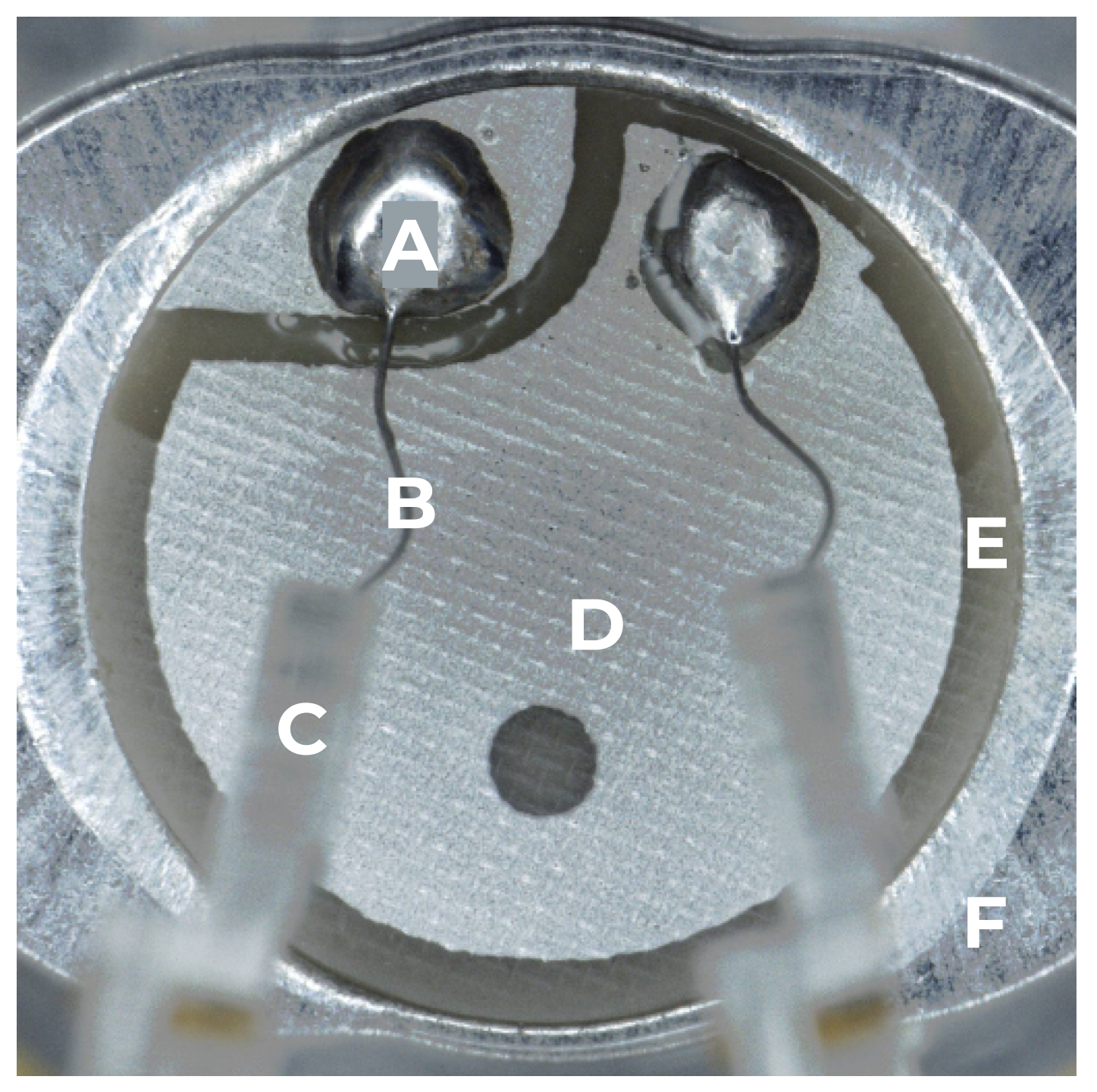}
    \caption{Product (VAD). A - soldering dot, B - wire, C - pin, D - piezo, E - piezo border, F - membrane.}
    \label{fig:product}
\end{figure}

\begin{longtable}[htp]{l | l}
      Examples & Description\\ 
      \midrule
      \raisebox{-0.5\totalheight}{\includegraphics[width=0.33\textwidth]{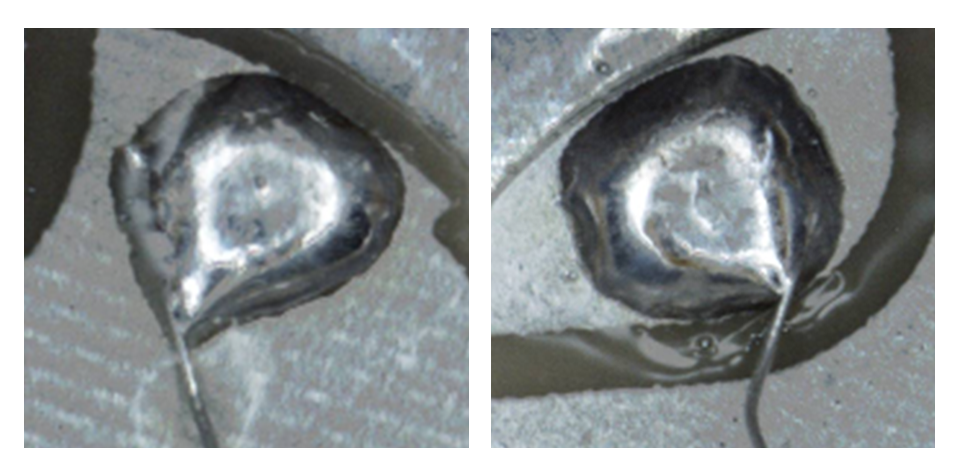}}
      & 
      The wire is too close to the side of the soldering \\
      \midrule
            \raisebox{-0.5\totalheight}{\includegraphics[width=0.33\textwidth]{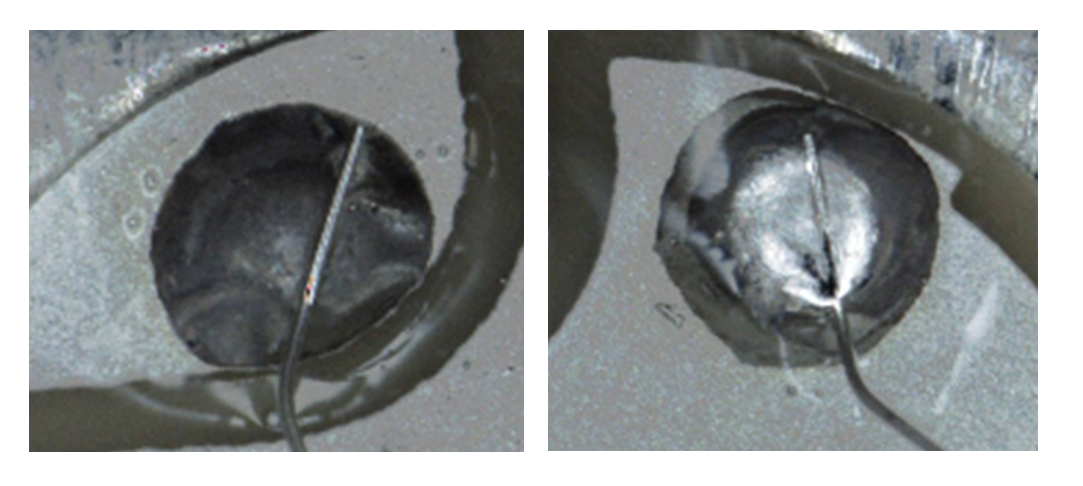}}
      & 
      The wire is on the top of the soldering \\
      \midrule
      \raisebox{-0.5\totalheight}{\includegraphics[width=0.33\textwidth]{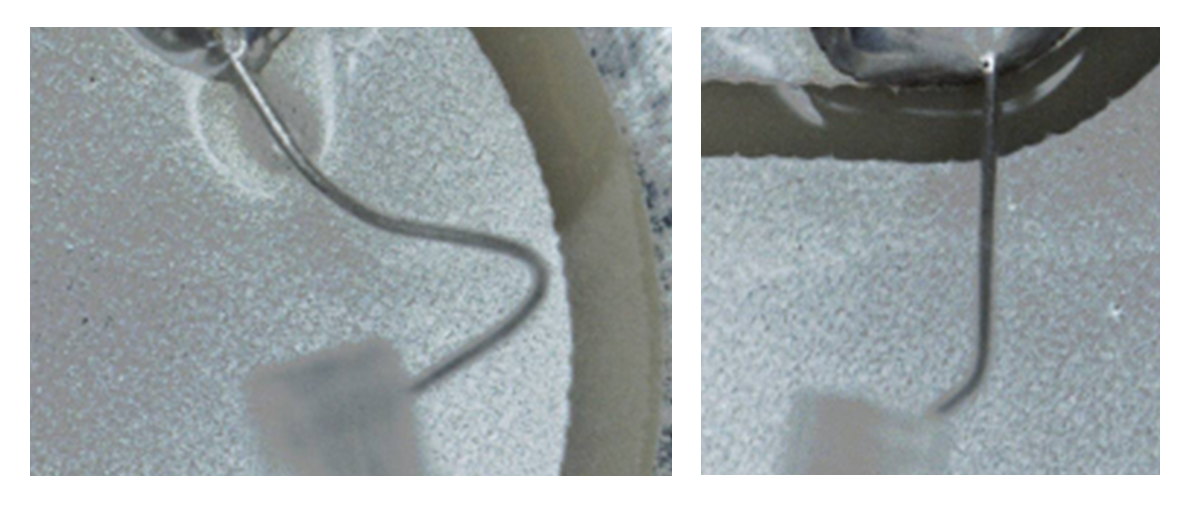}}
      & 
      The wire is poorly shaped, too bent or too straight\\
      \midrule
      \raisebox{-0.5\totalheight}{\includegraphics[width=0.33\textwidth]{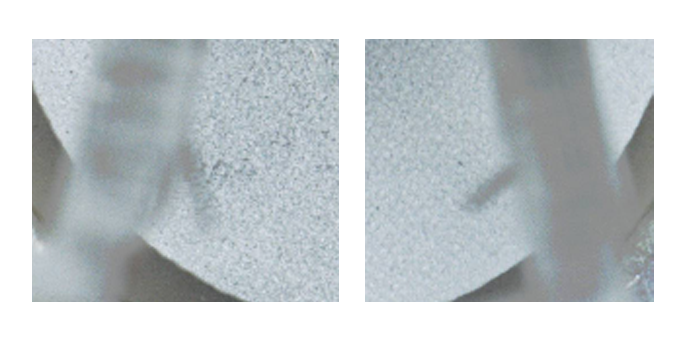}}
      & 
      The wire is not properly connected to the pin\\
      \midrule
      \raisebox{-0.5\totalheight}{\includegraphics[width=0.33\textwidth]{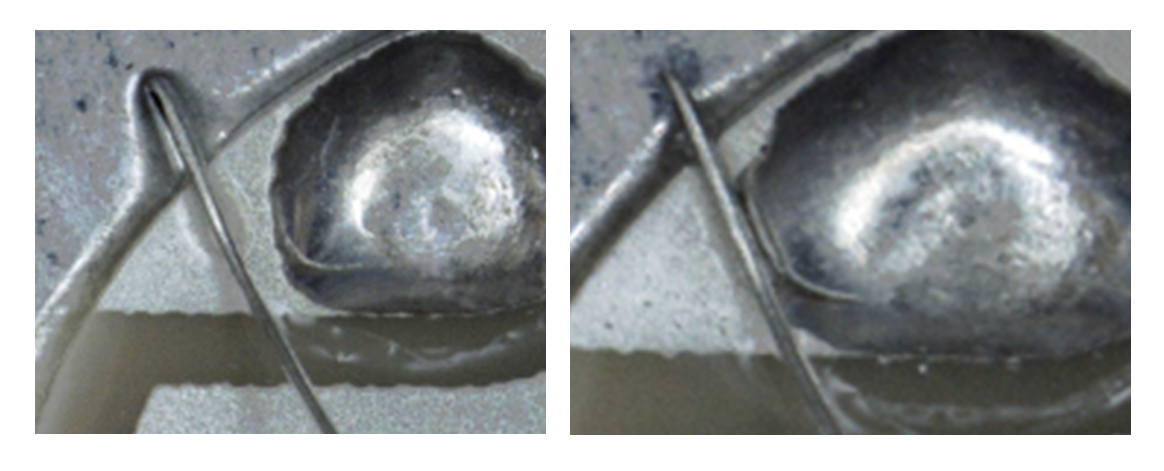}}
      & 
      The wire is out of soldering entirely\\
      \midrule
      \raisebox{-0.5\totalheight}{\includegraphics[width=0.33\textwidth]{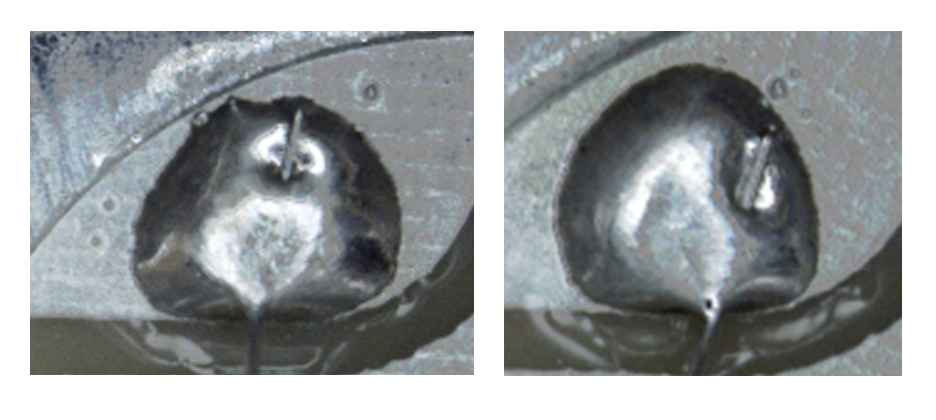}}
      & 
      The end of the wire is out \\
      \midrule
      \raisebox{-0.5\totalheight}{\includegraphics[width=0.33\textwidth]{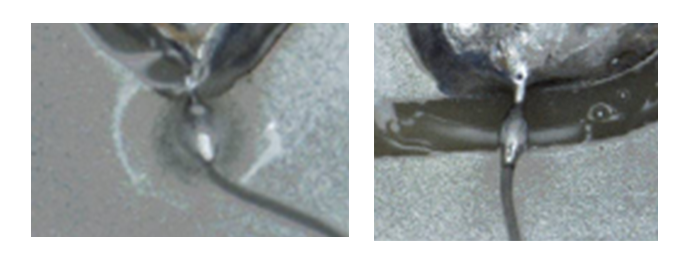}}
      & 
      Solder paste is on the wire \\
      \midrule
      \raisebox{-0.5\totalheight}{\includegraphics[width=0.33\textwidth]{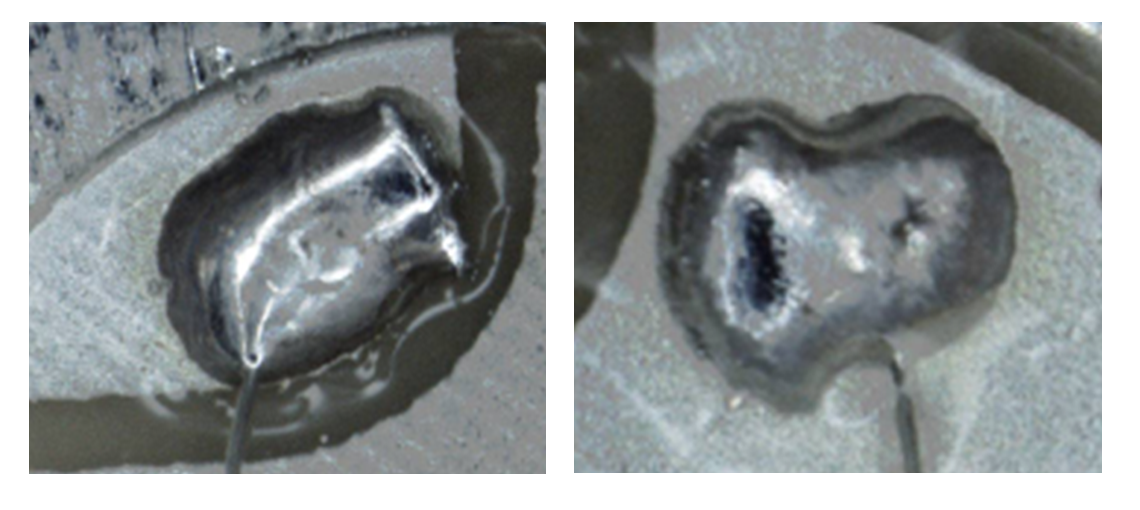}}
      & 
      Soldering is not complete\\
      \midrule
      \raisebox{-0.5\totalheight}{\includegraphics[width=0.33\textwidth]{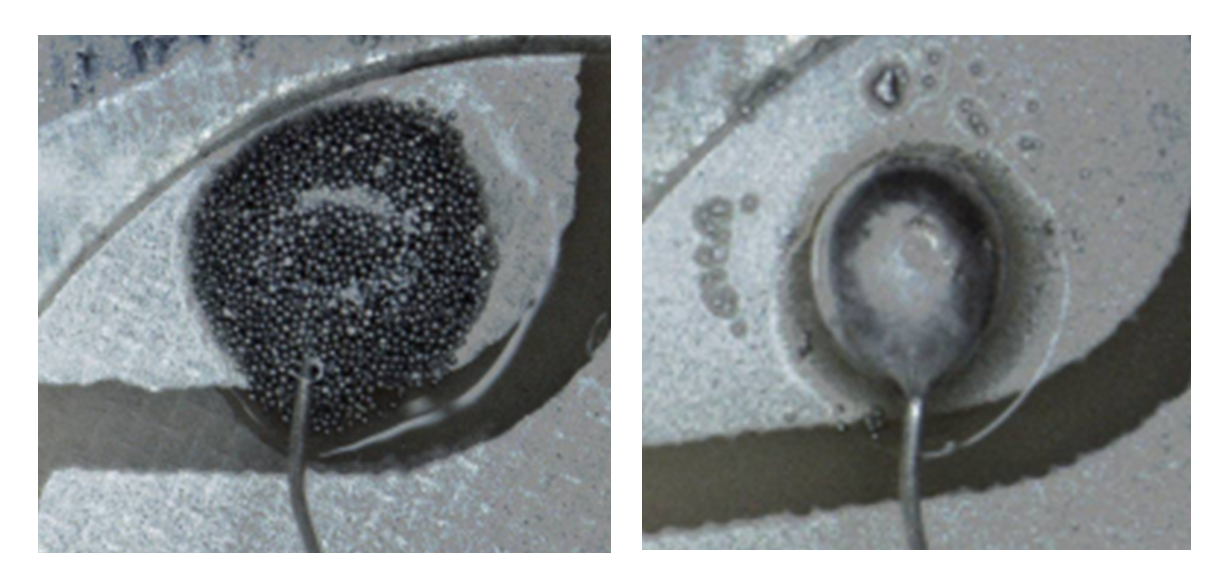}}
      & 
      Poor quality soldering\\
      \midrule
      \raisebox{-0.5\totalheight}{\includegraphics[width=0.33\textwidth]{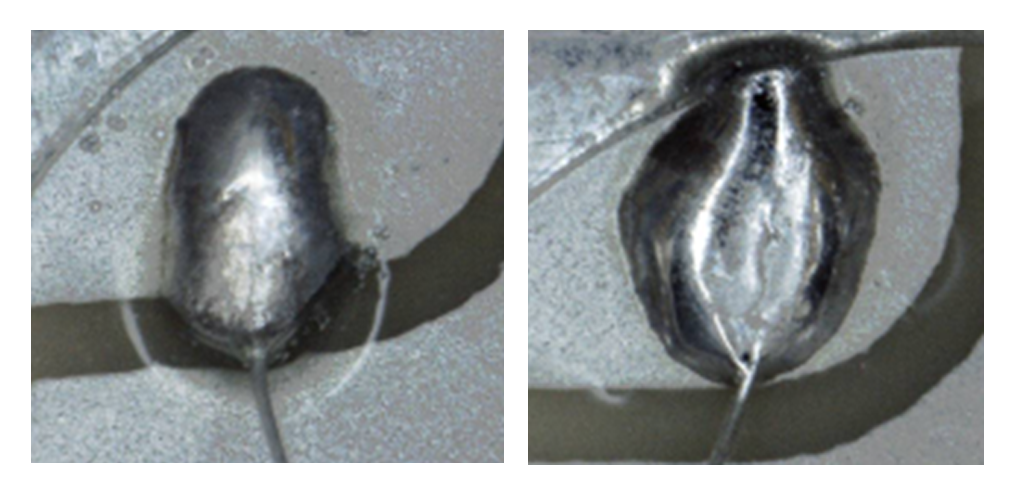}}
      & 
      Poorly shaped soldering \\
      \midrule
      \raisebox{-0.5\totalheight}{\includegraphics[width=0.33\textwidth]{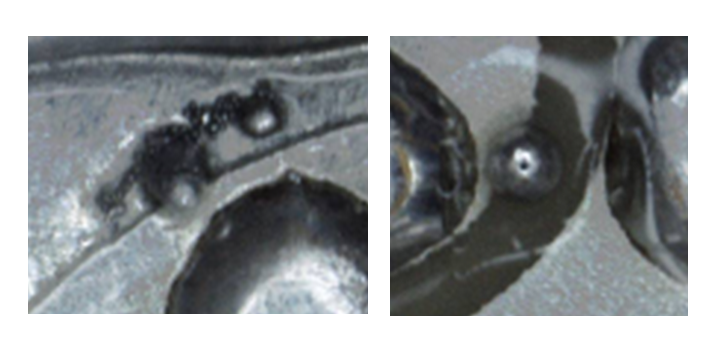}}
      & 
      Smaller solder balls pollution\\
      \midrule
      \raisebox{-0.5\totalheight}{\includegraphics[width=0.33\textwidth]{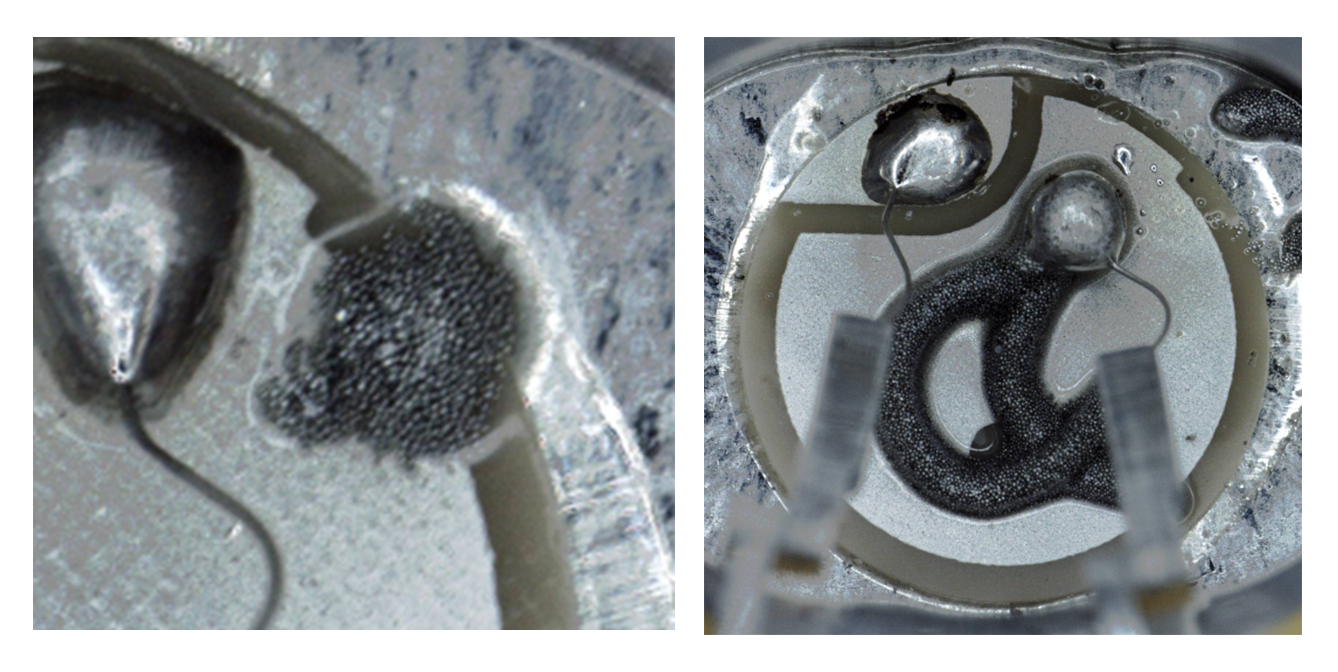}}
      & 
      Misplaced soldering paste\\
      \midrule
      \raisebox{-0.5\totalheight}{\includegraphics[width=0.33\textwidth]{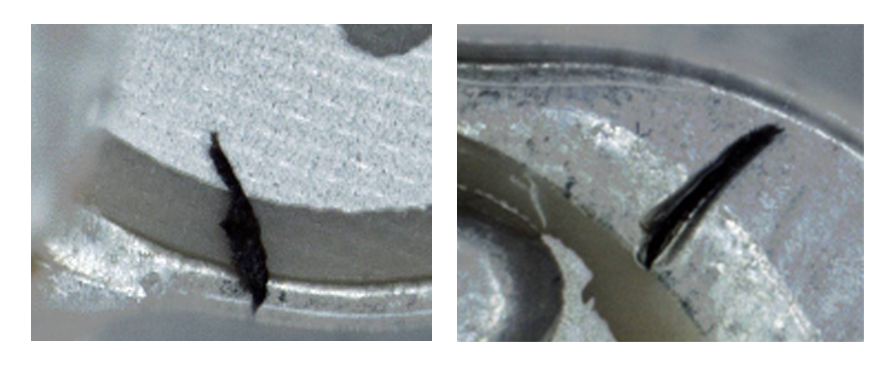}}
      & 
      Plastic pollution\\
      \midrule
      \raisebox{-0.5\totalheight}{\includegraphics[width=0.33\textwidth]{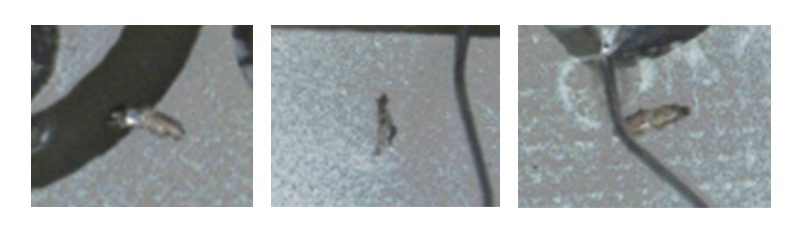}}
      & 
      Metal pollution \\
      \midrule
      \raisebox{-0.5\totalheight}{\includegraphics[width=0.33\textwidth]{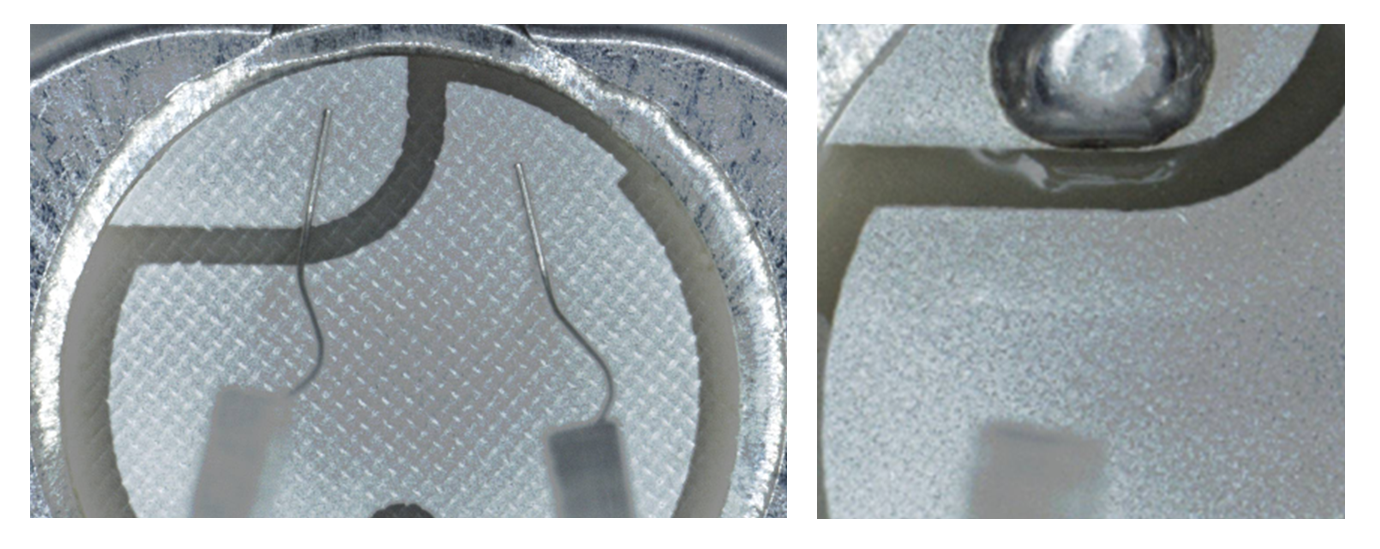}}
      & 
      Missing parts \\
      \midrule
      \raisebox{-0.5\totalheight}{\includegraphics[width=0.33\textwidth]{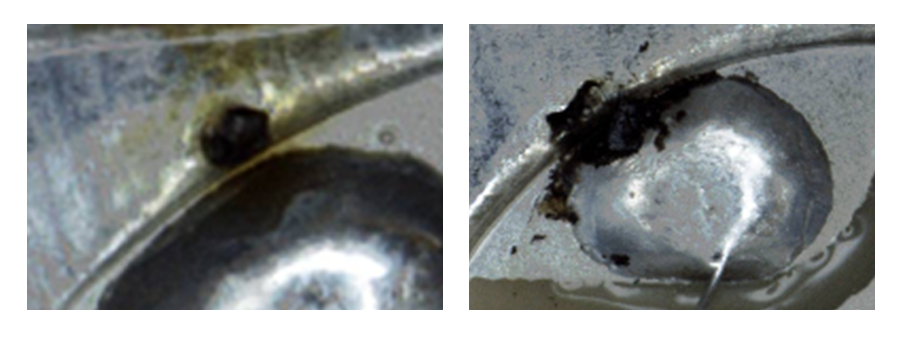}}
      & 
      Burned soldering or wires\\
      \midrule
      \raisebox{-0.5\totalheight}{\includegraphics[width=0.33\textwidth]{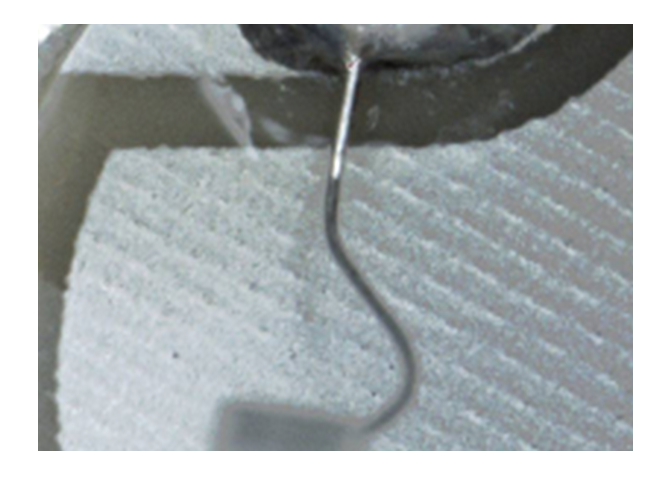}}
      & 
      The upper part of the wire is too long\\
      \midrule
      \raisebox{-0.5\totalheight}{\includegraphics[width=0.33\textwidth]{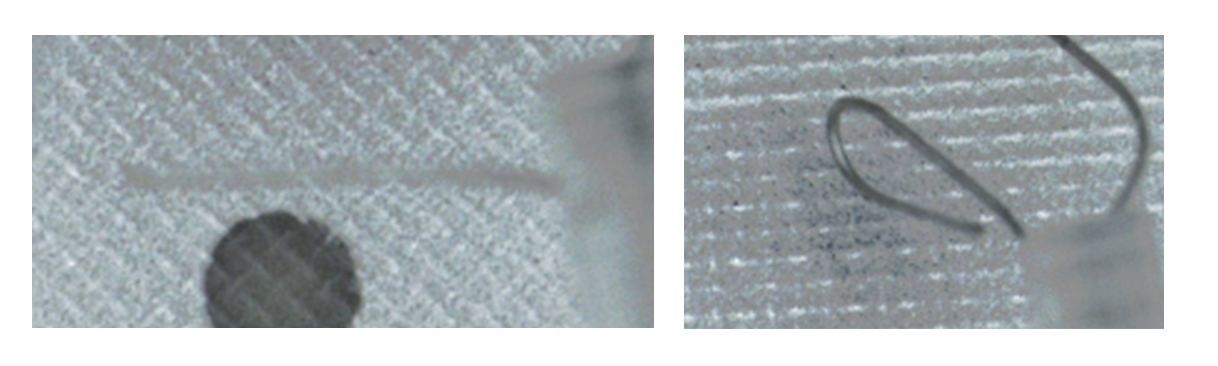}}
      & 
      Additional wires \\
      \midrule
      \raisebox{-0.5\totalheight}{\includegraphics[width=0.33\textwidth]{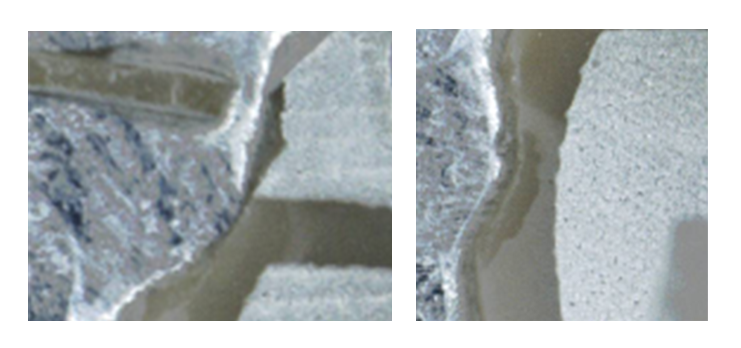}}
      & 
      Piezo is broken \\
      \midrule
      \raisebox{-0.5\totalheight}{\includegraphics[width=0.33\textwidth]{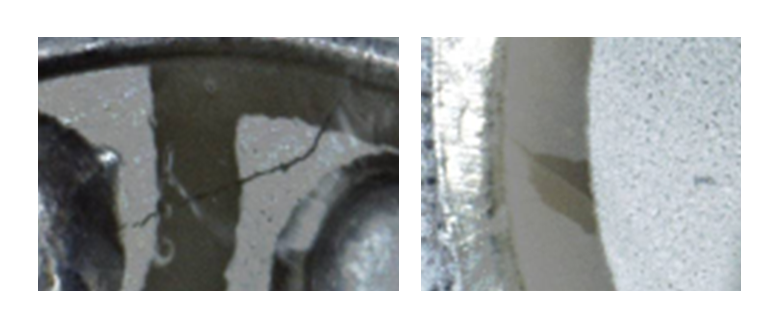}}
      & 
      Cracks in piezo\\
      \midrule
      \raisebox{-0.5\totalheight}{\includegraphics[width=0.33\textwidth]{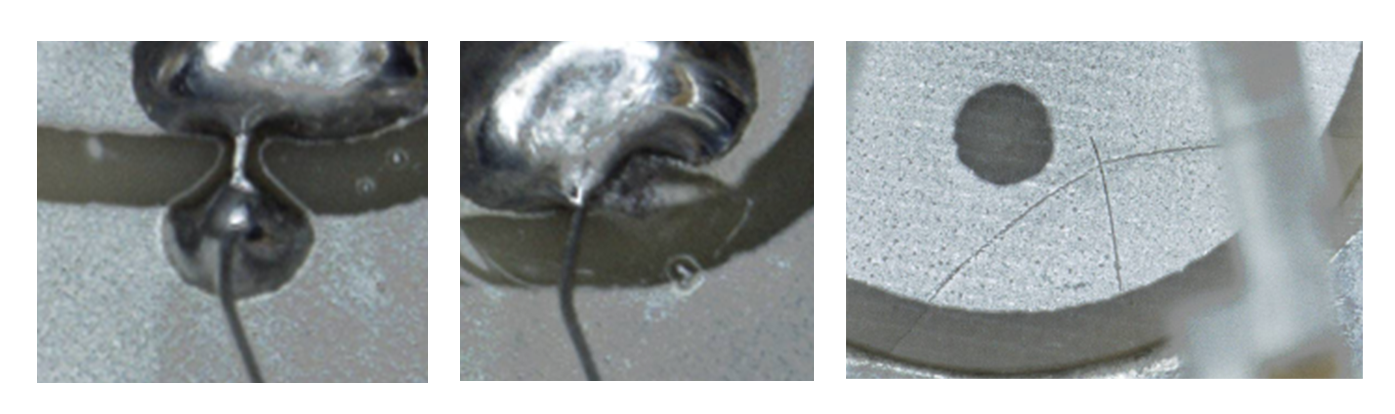}}
      & 
      Various rare defects\\
      \midrule
%      \end{tabularx}
\caption{Defects in VAD.}
\label{tbl:defects_vad2}
\end{longtable}

%--------------------------------------------------------------------------------
\section{Detailed results for VisA}
\label{appendix:visa}
In this section, we put the results for our supervised benchmark for VisA per class. It is worth noticing that the VisA paper \cite{zou2022spot} already has two supervised benchmarks, which were not widely adopted by the anomaly detection community. Their first supervised benchmark includes 60\% of all images, good and bad, used for training and the rest for testing. VisA paper already reports results of 99.7 Cl. AUROC with the supervised classifier used together with their method SPD. Such a high result for a supervised method can be explained by the fact that VisA contains a low number of anomaly types per class (4.7 on average), so 60 bad images per class for training were sufficient to recognize the rest. Their second supervised benchmark uses just 10 good and 10 bad images, and it also has not become too popular because anomaly detection methods usually try to utilize many good images, which are much easier to get than bad for real-world problems. There are very few methods that can tackle such problems successfully, such as PatchCore \cite{patchcore}. Our supervised benchmark for VisA is an attempt to fix these problems. We have both a high number of good parts, as well as 10 randomly selected bad parts for training, in a similar way to the popular supervised MVTec AD benchmark \footnote{\url{https://paperswithcode.com/sota/supervised-anomaly-detection-on-mvtec-ad}}. Results per class are shown in Table \ref{tab:visa_detailed}.

\bigskip \bigskip 

\begin{table}[htp]
\centering
\begin{tabularx}{490pt}{X | cccccccccccc | c}
Method & 
\begin{rotate}{50} candle \end{rotate} & 
\begin{rotate}{50} capsules \end{rotate} & 
\begin{rotate}{50} cashew \end{rotate} & 
\begin{rotate}{50} chewinggum \end{rotate} & 
\begin{rotate}{50} fryum \end{rotate} & 
\begin{rotate}{50} macaroni1 \end{rotate} & 
\begin{rotate}{50} macaroni2 \end{rotate} & 
\begin{rotate}{50} pcb1 \end{rotate} & 
\begin{rotate}{50} pcb2 \end{rotate} & 
\begin{rotate}{50} pcb3 \end{rotate} & 
\begin{rotate}{50} pcb4 \end{rotate} &
\begin{rotate}{50} pipe\_fryum \end{rotate} & 
Mean \\
\midrule
\multicolumn{14}{c}{\textit{One-class Anomaly Detection}} \\
\midrule
RD4AD & 94.6 & 85.4 & 96.3 & 98.8 & 93.7 & 96.5 & 84.7 & 96.0 & 97.1 & 95.9 & \textbf{99.9} & 97.5 & 94.7\\
EfficientAD & 98.3 & \textbf{92.3} & 96.0 & \textbf{100.0} & 98.5 & 99.1 & 96.7 & 98.6 & \textbf{99.9} & \textbf{98.4} & 98.3 & \textbf{99.4} & 97.9\\
\midrule
\multicolumn{14}{c}{\textit{Supervised Anomaly Detection}} \\
\midrule
DRA & 96.0 & 70.6 & 94.6 & 91.4 & \textbf{99.2} & 92.2 & 74.4 & 92.1 & 91.6 & 77.7 & 90.9 & 96.5 & 88.9\\
DevNet & 94.7 & 78.2 & 97.4 & 91.9 & 98.3 & 92.1 & 72.3 & 94.4 & 89.5 & 73.4 & 95.7 & 94.3 & 89.3\\
\midrule
\multicolumn{14}{c}{\textit{SegAD (ours)}} \\
\midrule
RD4AD + Ours & 98.4 & 80.1 & 98.9 & 99.3 & 96.2 & 97.4 & 90.6 & 96.4 & 96.3 & 94.1 & \textbf{99.9} & 95.8 & 95.3\\
EfficientAD + Ours & 98.7 & 89.7 & 98.6 & 99.9 & 98.6 & 99.5 & 98.1 & \textbf{99.5} & 99.7 & \textbf{98.4} & 99.3 & 99.2 & 98.3\\
All AD + Ours & \textbf{99.1} & 90.1 & \textbf{99.1} & 99.9 & 98.6 & \textbf{99.6} & \textbf{98.4} & 99.2 & \textbf{99.9} & 98.1 & 99.8 & \textbf{99.4} & \textbf{98.4}\\
\end{tabularx}
\caption{Results on VisA dataset, Cl. AUROC values for different classes. The best results are shown in bold. All AD means RD4AD and EfficientAD.
\label{tab:visa_detailed}}
\end{table}

Our method falls short with one class, capsules. This might be explained by some of the bad images having dark spots in the background, which are not defects, but together with image-level labels they provide SegAD with misleading information about the importance of the background. In any case, it shows that our method can sometimes make the results worse if the training data has semantic differences from the testing data. 
The supervised method from the original VisA high-shot supervised benchmark, mentioned in their paper, also shows lower results on the capsules class, 97.2 Cl. AUROC compared to the average result of 99.7, which might also indicate a difference between test and training data.

%--------------------------------------------------------------------------------
\section{Additional experiments with MVTec AD}
\label{appendix:mvtec}
This section shows the results of additional experiments with MVTec AD per class. The task is to investigate how our method will work on another dataset, it is not a benchmark. 20 random bad images from the test set are selected for training SegAD over ten different seeds for the selection process. 10\% of good images from the training dataset are used to train SegAD as well.
MVTec AD includes 15 classes, 5 of which are textures and 10 are simple objects~\cite{mvtec}. Training datasets consist of 242 good images per class on average (compared with 720 for VisA and 2000 for VAD), which leaves us with fewer good images for training SegAD. EfficientAD + Ours shows worse results compared to EfficientAD. This displays that some anomaly detectors might be not working with our method in some cases for reasons that require further investigation. Nevertheless, other anomaly detectors show improvement.

\bigskip \bigskip

\begin{table}[htp]
\small
\centering
\begin{tabularx}{500pt}{X | ccccccccccccccc | c}
M & 
\begin{rotate}{50} bottle \end{rotate} & 
\begin{rotate}{50} cable \end{rotate} & 
\begin{rotate}{50} capsule \end{rotate} & 
\begin{rotate}{50} carpet \end{rotate} & 
\begin{rotate}{50} grid \end{rotate} & 
\begin{rotate}{50} hazelnut \end{rotate} & 
\begin{rotate}{50} leather \end{rotate} & 
\begin{rotate}{50} metal\_nut \end{rotate} & 
\begin{rotate}{50} pill \end{rotate} & 
\begin{rotate}{50} screw \end{rotate} & 
\begin{rotate}{50} tile \end{rotate} &
\begin{rotate}{50} toothbrush \end{rotate} & 
\begin{rotate}{50} transistor \end{rotate} & 
\begin{rotate}{50} wood \end{rotate} &
\begin{rotate}{50} zipper \end{rotate} & 
Mean \\
\midrule
\multicolumn{15}{c}{\textit{One-class Anomaly Detection}} \\
\midrule
F & \textbf{100.0} & 88.2 & 95.0 & 99.2 & \textbf{100.0} & 78.4 & 99.6 & 97.6 & 94.3 & 74.7 & \textbf{100.0} & 88.1 & 89.8 & 99.4 & 97.7 & 93.5\\
R & 99.8 & 94.7 & 97.5 & 99.2 & 95.3 & \textbf{100.0} & \textbf{100.0} & \textbf{100.0} & 96.2 & \textbf{98.4} & \textbf{100.0} & 93.8 & 97.1 & \textbf{99.6} & 97.9 & 98.0\\
E & \textbf{100.0} & 91.1 & 97.6 & 97.1 & 98.2 & 99.6 & 99.4 & 99.9 & 97.9 & 92.4 & 99.7 & \textbf{99.8} & \textbf{99.8} & \textbf{99.6} & 98.8 & 98.1\\
\midrule
\multicolumn{15}{c}{\textit{SegAD (ours)}} \\
\midrule
F+O & 99.7 & 71.6 & 94.4 & 96.9 & 97.5 & 98.2 & 99.4 & 96.6 & 95.2 & 74.9 & 99.9 & 99.2 & 95.4 & 98.5 & \textbf{99.4} & 94.5\\
R+O & 99.6 & \textbf{95.7} & 97.9 & \textbf{98.6} & 99.7 & \textbf{100.0} & \textbf{100.0} & \textbf{100.0} & 97.5 & 95.7 & \textbf{100.0} & 97.3 & 95.9 & 99.0 & 98.4 & \textbf{98.4}\\
E+O & 99.5 & 88.6 & 95.9 & 95.6 & 98.1 & 99.5 & 98.5 & 99.2 & 97.3 & 90.4 & 99.9 & 95.2 & 96.7 & 99.1 & 96.3 & 96.7\\
A+O & \textbf{100.0} & 92.5 & \textbf{98.3} & 98.5 & 99.6 & \textbf{100.0} & \textbf{100.0} & 99.5 & \textbf{98.2} & 95.2 & \textbf{100.0} & 96.3 & 99.3 & 99.0 & 99.3 & \textbf{98.4}\\
\end{tabularx}
\caption{Results on MVTec AD dataset, Cl. AUROC values for different classes. The best results are shown in bold. M = \textit{Method}, F = \textit{FastFlow}, R = \textit{RD4AD}, E = \textit{EfficientAD}, O = \textit{Ours}, A = \textit{All AD} (includes FastFlow, RD4AD, and EfficientAD).
\label{tab:mvtec_detailed}}
\end{table}

\label{sec:appendices}

\end{document}